\documentclass[lettersize,journal]{IEEEtran}
\usepackage{amsmath,amsfonts}
\usepackage{amssymb}
\usepackage{algorithmic}
\usepackage{algorithm}
\usepackage{array}
\usepackage[caption=false,font=normalsize,labelfont=sf,textfont=sf]{subfig}
\usepackage{textcomp}
\usepackage{stfloats}
\usepackage{url}
\usepackage{verbatim}
\usepackage{graphicx}
\usepackage{cite}
\usepackage{adjustbox}
\usepackage{multirow}
\usepackage{booktabs}
\usepackage{xcolor}
\usepackage{bbding}
\usepackage{pifont}
\usepackage{hyperref}
\usepackage{colortbl}
\usepackage{threeparttable}
\usepackage{makecell}

\begin{document}

% \title{YOLO-LLTS: Real-Time Traffic Sign Detection with Prior Knowledge and Multi-Scale Features Fusion in Low-Light Environments}

\title{YOLO-LLTS: Real-Time Low-Light Traffic Sign Detection via Prior-Guided Enhancement and Multibranch Feature Interaction}
\author{Ziyu Lin, Yunfan Wu, Yuhang Ma, Junzhou Chen, Ronghui Zhang, 
Jiaming Wu, Guodong Yin,\\ and Liang Lin,~\IEEEmembership{Fellow, IEEE}
        % <-this % stops a space
\thanks{
This project is jointly supported by the National Natural Science Foundation of China (Nos.52172350, W2421069 and 51775565), the Tongchuang Intelligent Medical Inter-disciplinary Talent Training Fund of Sun Yat-sen University (No.76160-54990001), the Guangdong Basic and Applied Research Foundation (No. 2022B1515120072), the Guangzhou Science and Technology Plan Project (No.2024B01W0079), the Nansha Key RD Program (No.2022ZD014), the Science and Technology Planning Project of Guangdong Province (No.2023B1212060029). (Corresponding author: Ronghui Zhang.)

Ziyu Lin, Yunfan Wu, Yuhang Ma, Junzhou Chen, Ronghui Zhang are with the Guangdong Key Laboratory of Intelligent Transportation System, School of intelligent systems engineering, Sun Yat-sen University, Guangzhou 510275, China (e-mail: linzy88@mail2.sysu.edu.cn, wuyf227@mail2.sysu.edu.cn, mayh39@mail2.sysu.edu.cn, chenjunzhou@mail.sysu.edu.cn, zhangrh25@mail.sysu.edu.cn).

Jiaming Wu is with the Department of Architecture and Civil Engineering, Chalmers University of Technology, Sven Hultins gata 6 SE-412 96, Gothenburg, Sweden (jiaming.wu@chalmers.se).

Guodong Yin is with the School of Mechanical Engineering, Southeast University, Nanjing 211189, China (e-mail: ygd@seu.edu.cn).

Liang Lin is with the School of Computer Science and Engineering, Sun Yat-sen University, Guangzhou 510275, China (e-mail: linliang@ieee.org).
}% <-this % stops a space
}

\maketitle

\begin{abstract}
Traffic sign detection is essential for autonomous driving and Advanced Driver Assistance Systems (ADAS). However, existing methods struggle to address the challenges of poor image quality and insufficient information under low-light conditions, leading to a decline in detection accuracy and affecting driving safety. To address this issue, we propose YOLO-LLTS, an end-to-end real-time traffic sign detection algorithm specifically designed for low-light environments. YOLO-LLTS introduces three main contributions: the HRFM-SOD module retains more information about distant or tiny traffic signs compared to traditional methods; the MFIA module interacts features with different receptive fields to improve information utilization; the PGFE module enhances detection accuracy by improving brightness, edges, contrast, and supplementing detail information. Additionally, we construct a new dataset, the Chinese Nighttime Traffic Sign Sample Set (CNTSSS), covering diverse nighttime scenarios. Experiments show that YOLO-LLTS achieves state-of-the-art performance, outperforming previous best methods by 2.7\% mAP50 and 1.6\% mAP50:95 on TT100K-night, 1.3\% mAP50 and 1.9\% mAP50:95 on CNTSSS, 7.5\% mAP50 and 9.8\% mAP50:95 on GTSDB-night, and superior results on CCTSDB2021. Deployment on edge devices confirms its real-time applicability and effectiveness. The code and the dataset are available at https://github.com/linzy88/YOLO-LLTS.

\end{abstract}

\vspace{-7pt}
\section{Introduction}
% \vspace{-1pt}
\IEEEPARstart{T}{raffic} sign detection plays a crucial role in advanced driver-assistance systems (ADAS) and autonomous driving perception technology. ADAS systems and autonomous vehicles equipped with traffic sign recognition algorithms are able to obtain real-time traffic regulations and road information, thereby making accurate driving decisions to ensure road safety. Existing object detection methods have demonstrated strong capabilities in accurately detecting various traffic elements, including pedestrians \cite{pedestrians1,pedestrians2,pedestrians3,MFFSODNet}, vehicles \cite{vehicle1,vehicle2,vehicle3,YOLOv8-QSD,MSFFA-YOLO}, and traffic lights \cite{trafficlights1,trafficlights2,trafficlights3}. However, the detection of traffic signs remains a considerable challenge, primarily due to their small size and the complexity involved in distinguishing them from other objects in the scene, making this task particularly difficult in real-world scenarios. In high-resolution images with dimensions of 2048$\times$2048 pixels, a sign may occupy only a small area of approximately 30$\times$30 pixels. Due to the extremely low resolution and limited information, significant efforts have been made in recent years to enhance the performance of small object detection. Existing traffic sign detection algorithms \cite{chen2022,23,25} have been improved to address the characteristics of small targets, enabling effective detection of traffic signs during the daytime. 

\begin{figure}[t]
    \centering
    \includegraphics[width=0.48\textwidth]{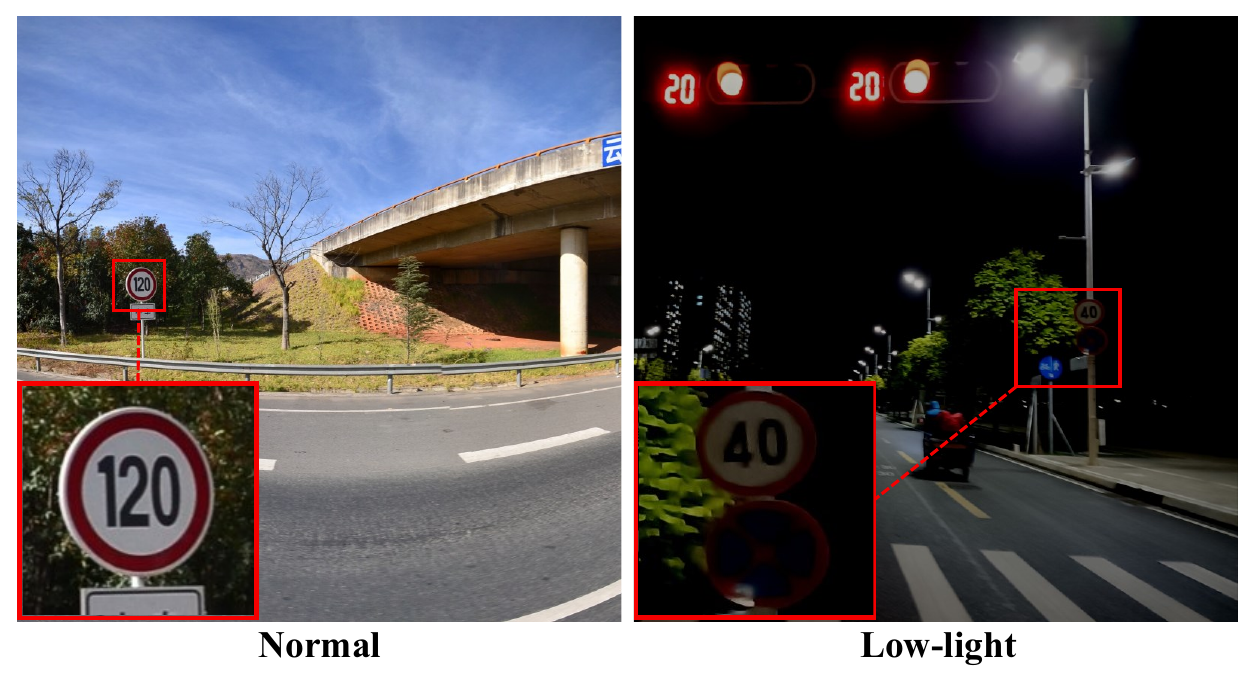} % 调整宽度为栏宽的一半左右
    \caption{The comparison of traffic signs in normal and low-light environments. The edges of traffic signs under low light are unclear and detecting the targets becomes challenging.}
    \label{intro}
\end{figure}

However, with the increasing number of low-light traffic accidents and the growing demand for all-weather systems, the importance of traffic sign recognition under low-light conditions has garnered more attention. As shown in Fig. \ref{intro}, visibility decreases and image noise increases, complicating the driving scenario in low-light environments. Facing the dual challenges of low-resolution small objects and low-visibility low-light conditions, existing methods struggle to clearly capture the features of traffic signs for detection and classification. 

\begin{figure*}[htb]
    \centering
    \includegraphics[width=\textwidth]{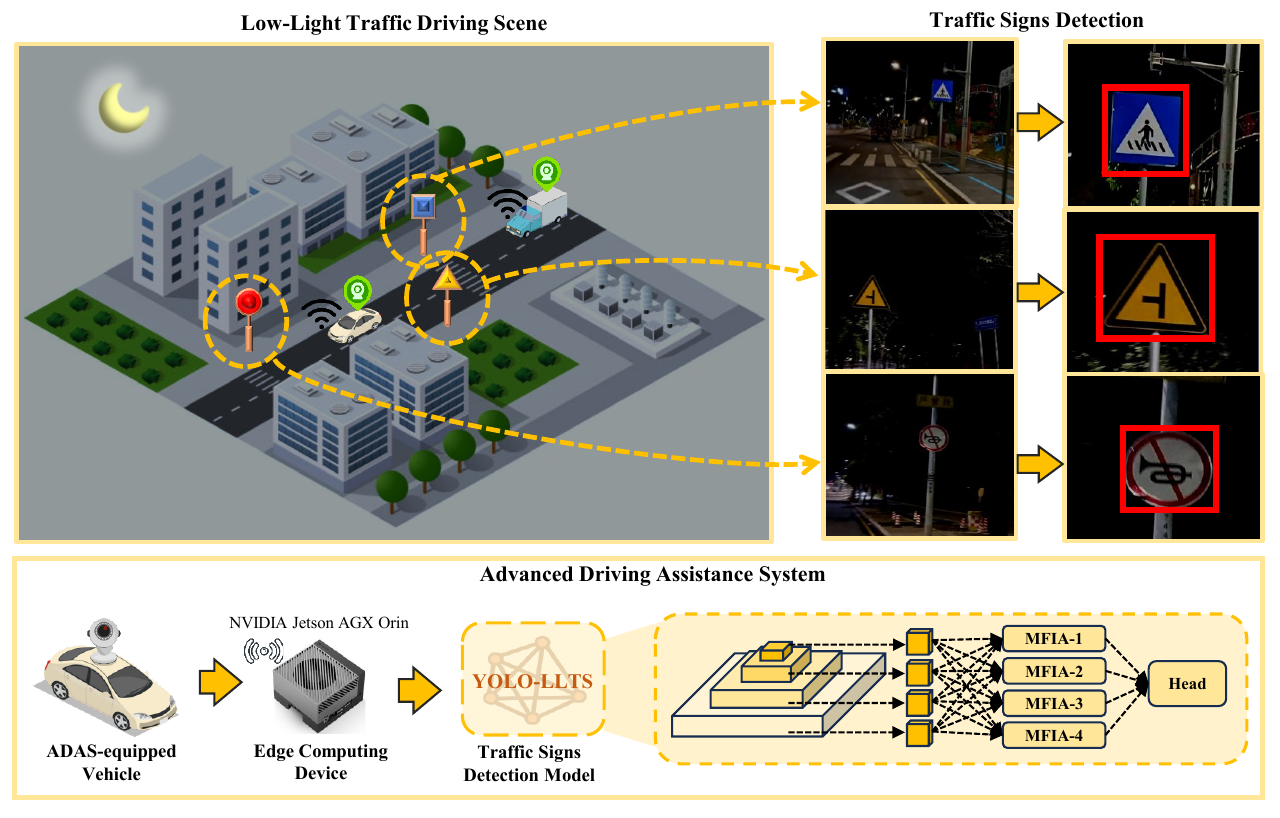}
    \caption{Application Scenarios of Traffic Sign Detection in advanced driver-assistance systems \cite{ico,52112,iconfinder}. Vehicles equipped with ADAS systems capture real-time image information through cameras. By detecting nighttime traffic signs using YOLO-LLTS, the system alerts the driver to take corrective actions or directly controls the vehicle to enhance driving safety.}
    \label{first}
\end{figure*}

A straightforward approach is to pre-enhance the images using advanced low-light enhancement techniques and then apply object detection algorithms to identify the enhanced images. The approach of addressing image enhancement and object detection as discrete tasks frequently results in a challenge of compatibility. Moreover, simply concatenating the two models results in slow inference speeds, which fails to meet the real-time requirements of ADAS. Chowdhury et al. \cite{Chowdhury2019} utilized optimal reinforcement learning strategies and various Generative Adversarial Network (GAN) \cite{12} models to enhance the training data for traffic sign recognition. However, this approach is heavily reliant on data labels, making it suitable only for specific datasets. Zhang et al. \cite{Zhang2024} enhanced model robustness through exposure, tone, and brightness filters, enabling end-to-end training. However, simply brightening low-light images led to the loss of original information and introduced more noise, which undermined the original intent. Sun et al. \cite{llth2024} proposed LLTH-YOLOv5, which enhanced images using pixel-level adjustments and non-reference loss functions. However, this method introduced an enhancement-specific loss function based on YOLO, meaning that end-to-end training was not fully achieved.

To address the dual challenges of small targets with low resolution and low visibility in low-light environments, we have designed an end-to-end traffic sign recognition algorithm specifically for low-light conditions. As shown in Fig. \ref{first}, low-light conditions can impair the driver's visibility, leading to the neglect of traffic signs and potential violations of traffic regulations. The ADAS, equipped with our YOLO-LLTS model, alerts the driver to take corrective actions or directly controls the vehicle upon detecting traffic signs. For example, when the system detects a speed limit sign, it automatically adjusts the vehicle's speed to comply with the road's speed limit. Similarly, if a stop sign is detected, the system either alerts the driver or automatically applies the brakes. YOLO-LLTS improved YOLOv8 by utilizing high-resolution images to extract clearer features, and it also designed a new Multibranch features interaction attention module to fuse features from different receptive fields. Additionally, we have developed a prior-providing module that not only enhances the image but also supplements its details. Our algorithm effectively mitigates the poor performance of existing methods under low-light conditions, thereby improving the safety of autonomous driving systems. Furthermore, to address the lack of low-light scene data in existing traffic sign datasets, we have constructed a multi-scene Chinese traffic sign dataset ranging from dusk to deep night, providing a foundational resource for industry research in this area. 

In summary, the main contributions of our proposed method are as follows:
\begin{itemize}
\item [1)] \textbf{Chinese Nighttime Traffic Sign Sample Set:} To address the lack of traffic sign datasets under low-light conditions, we constructed a novel dataset named CNTSSS. This dataset was collected across 17 cities in China, containing images captured under various nighttime lighting conditions ranging from dusk to deep night. It covers diverse scenarios, including urban, highway, and rural environments, as well as clear and rainy weather conditions.

\item [2)] \textbf{High-Resolution Feature Map for Small Object Detection (HRFM-SOD):} To address the challenge of indistinct features for small objects in low-light conditions, we propose the HRFM-SOD module, which utilizes high-resolution feature maps for detection. This module effectively alleviates the feature dilution problem encountered in traditional PANet frameworks when detecting small targets, thus improving both detection accuracy and inference speed.

\item [3)] \textbf{Multibranch Feature Interaction Attention Module (MFIA):} To enhance the model's capability of capturing information from multiple receptive fields, we introduce the MFIA module. This module facilitates deep interaction and fusion of multi-scale features across both channel and spatial dimensions. Unlike previous methods that focus solely on single-scale attention mechanisms, MFIA effectively integrates multi-scale, semantically diverse features.

\item [4)] \textbf{Prior-Guided Feature Enhancement Module (PGFE):} To overcome common image quality issues such as noise, reduced contrast, and blurriness inherent in low-light environments, we propose the PGFE module. This module utilizes prior knowledge to enhance images and supplement image details, significantly boosting detection performance under challenging low-light conditions.
\end{itemize}

The remaining parts of this article are organized as follows. Section \ref{sec:related_work} reviews the related works on low-light image enhancement methods, target detection methods in complex scenarios, and small target detection methods. Section \ref{sec:methods} describes the details of the CNTSSS dataset we constructed and the low-light traffic sign detection algorithm YOLO-LLTS. Section \ref{sec:experiments} presents the experimental analysis and results. Finally, Section \ref{sec:conclusion} concludes the paper and discusses future work.

\section{Related Works}
\label{sec:related_work}
For the task of detecting traffic signs, the greatest challenge lies in the small size of the traffic signs and the precise detection and localization of the traffic signs in various complex scenarios. Therefore, we systematically review the relevant research status in three aspects: low-light image enhancement (LLIE) methods, target detection methods in complex scenarios, and small target detection methods.
\subsection{Low-Light Image Enhancement Methods (LLIE Methods)}
LLIE methods are able to effectively enhance the quality of images under low-light conditions. Currently, the enhancement methods in this field mainly consist of two classes: traditional methods and machine learning methods.

Traditional methods in LLIE primarily emphasized histogram equalization and strategies derived from Retinex theory \cite{1}. Histogram equalization enhanced image brightness by expanding the dynamic range of pixel values, which included global methods \cite{2} and local methods \cite{3}. Methods based on Retinex decomposed the image into illumination and reflection components, assuming that the reflection component remained consistent under different illumination conditions. For example, Fu et al. \cite{4,5} first used two norm-constrained illumination and proposed a two norm-based optimization solution, while the Retinex model proposed by Li et al. \cite{6} considering noise, estimating the illumination map by solving an optimization problem. However, these traditional methods usually rely on manually extracted features and may struggle to achieve ideal enhancement effects under complex lighting conditions.

LLIE based on Machine learning methods mainly consist of two categories: supervised learning and unsupervised learning. Supervised learning approaches generally rely on extensive datasets of low-light images as well as their paired images under normal-light conditions to facilitate effective training. For example, LLNet \cite{7}, as the first deep learning method for low light image enhancement (LLIE) that introduced an end-to-end network, was trained on simulated data with random gamma correction. Furthermore, Wei et al. \cite{8} creatively combined Retinex theory with convolutional neural networks, dividing the network into modules of decomposition, adjustment and reconstruction, and trained it using their self-built LOL dataset. The performance of these methods relies heavily on the quality and diversity of the paired training datasets. Unsupervised learning methods focus on enhancing low-light images without paired training data. For example, Zero-DCE \cite{9} approached image enhancement as a task of estimating image-specific curves using a deep network, driving the network's learning process through a series of meticulously crafted loss functions to achieve enhancement without the need for paired data. EnlightenGAN \cite{10} used an attention mechanism-based U-Net \cite{11} as a generator and employed the method of GAN to perform image enhancement tasks without paired training data. Cui et al. \cite{15} proposed the Illumination Adaptive Transformer (IAT) model, which uses attention mechanisms to adjust parameters related to the image signal processor (ISP), effectively enhancing targets under various lighting conditions. These methods demonstrate the potential for unsupervised learning in LLIE and demonstrate the flexibility and adaptability of deep learning models to diverse lighting conditions.

\subsection{Small Target Detection Methods}
Detecting small targets in object detection is a challenging task. Small targets are often plagued by low resolution, and their feature extraction and precise detection are extremely difficult due to interference from various background information. Moreover, since the positions of small target detection objects are usually not fixed and may appear anywhere in the image, including peripheral areas or overlapping objects, their precise localization is even more challenging. Data augmentation, multi-receptive field learning, and context learning are effective strategies to boost the performance of small target detection.

Data augmentation, as a simple and effective strategy, can effectively enhance the ability to extract features of small target objects by increasing the diversity of the training set. Cui et al. \cite{17} pasted objects into different backgrounds to directly augment the rare classes in the dataset; Zhang et al. \cite{18} used GAN to perform data augmentation, optimizing the model's stability and robustness; Xie et al. \cite{19} increased the number of high-difficulty negative samples in data set balancing and data augmentation during training, expanding the existing dataset by simulating complex environmental changes.

Multi-scale fusion learning integrates deep semantic information with shallow representation information, effectively mitigating the fading of small object features and positional details in the detection network layer by layer. SODNet \cite{20} adaptively acquires corresponding spatial information through multi-scale fusion, thereby strengthening the network's capacity to extract features of small target objects. Ma et al. \cite{21} used deconvolution to upsample deep semantic information and then combined it with shallow representation information to build a feature pyramid, improving detection accuracy. TsingNet \cite{22} constructed a bidirectional attention feature pyramid, using both top-down and bottom-up subnets to perceive foreground features and reduce semantic gaps between multiple scales, effectively detecting small targets. MIAF-Net \cite{23}, consisting of a lightweight FCSP-Net backbone, an Attention-Balanced Feature Pyramid Network (ABFPN), and a Multi-Scale Information Fusion Detection Head (MIFH), can not only effectively extract features of small objects but also enhance the association between foreground features and contextual information through a self-attention mechanism, thereby performing well in small target detection tasks. 

The appearance features of small targets are usually not very prominent, so appropriate context modeling can improve the detector's performance on small target detection tasks. AGPCNet \cite{24} fused context information from multiple scales through a context pyramid module, achieving better feature representation of small targets. YOLO-TS \cite{25} optimized the receptive fields of multi-receptive field feature maps and performed multi-receptive field object detection on high-resolution feature maps rich in context information.

\subsection{Object Detection Methods in Complicated Scenarios}
Object detection in complicated scenarios imposes higher robustness requirements on detection models compared to conventional target detection, which typically includes tasks under various adverse weather conditions and different lighting conditions.

IA-yolo \cite{13} employs a differentiable image processing (DIP) module to adaptively enhance each image to improve the robustness of the model. PE-yolo \cite{14} utilizes a Pyramid Enhancement Network (PENet) to construct a low-light object detection framework and employs an end-to-end training method to simplify the training process, effectively completing target detection tasks under various low-light conditions. Yang et al. \cite{16} proposed a Dual-Mode Serial Night Road Object Detection Model (OMOT) based on depthwise separable convolution and a self-attention mechanism. OMOT significantly improves the detection accuracy of vehicles and pedestrians at night by leveraging a lightweight object proposal module and a classification module enhanced with self-attention mechanisms. The model not only considers vehicle light features, but also enhances the extraction of nighttime features through self-attention, demonstrating robust performance in complex environments.

The problem of traffic sign detection under low-light conditions is a subtask of object detection in complex scenes. It can be decomposed into two challenges: object detection in complex scenes and small object detection. There is limited existing research on this topic. Zhang et al. \cite{Zhang2024} enhanced images using exposure, tone, and luminance filters with a small convolutional network for predicting filter parameters during training. And they improved object detection accuracy with a Feature Encoder and an explicit objectness branch. Sun et al. \cite{llth2024} enhanced images using pixel-level adjustments and non-reference loss functions at the low-light enhancement step. The framework improved upon YOLOv5 by replacing PANet with BIFPN and introducing a transformer-based detection head for better small target detection during the object detection phase.

\begin{figure}[t]
    \centering
    \includegraphics[width=0.48\textwidth]{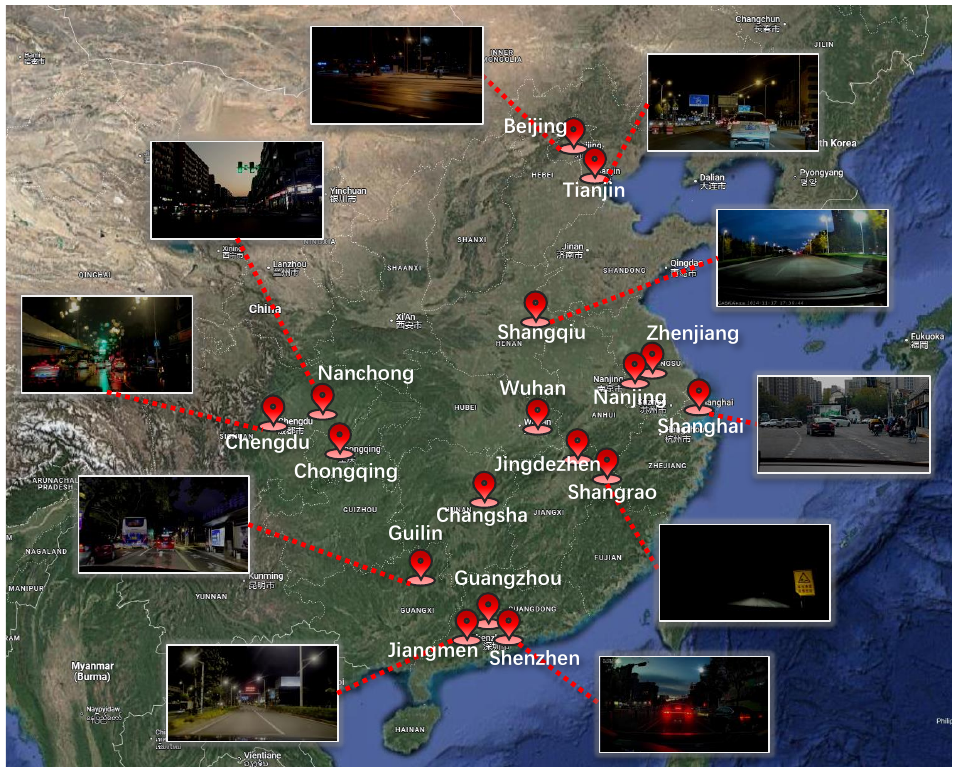} % 调整宽度为栏宽的一半左右
    \caption{Comprehensive collection of nighttime traffic sign images captured across 17 cities in China, including Beijing, Shanghai, Guangzhou, Shenzhen, Jiangmen, Chongqing, Chengdu, Nanchong, Wuhan, Changsha, Tianjin, Nanjing, Zhenjiang, Shangqiu, Shangrao, Guilin, and Jingdezhen.}
    \label{city}
\end{figure}

\section{Methods}
\label{sec:methods}
In this section, we provide a detailed introduction to our dataset and modules. Firstly, we describe the foundational aspects and detailed information of the dataset, which is intended to support further research in this area. Secondly, we introduce the overall architecture of the YOLO-LLTS model. Thirdly, We describe the design of the HRFM-SOD module, which alleviates the issue of blurred features of small objects in low-light conditions. Then, we explain the design of the MFIA module, which enhances the model’s ability to perform feature interactions across multiple receptive fields. Finally, we present the PGFE module, which utilizes prior knowledge to improve the quality of low-light images.

\begin{figure}[t]
    \centering
    \includegraphics[width=0.48\textwidth]{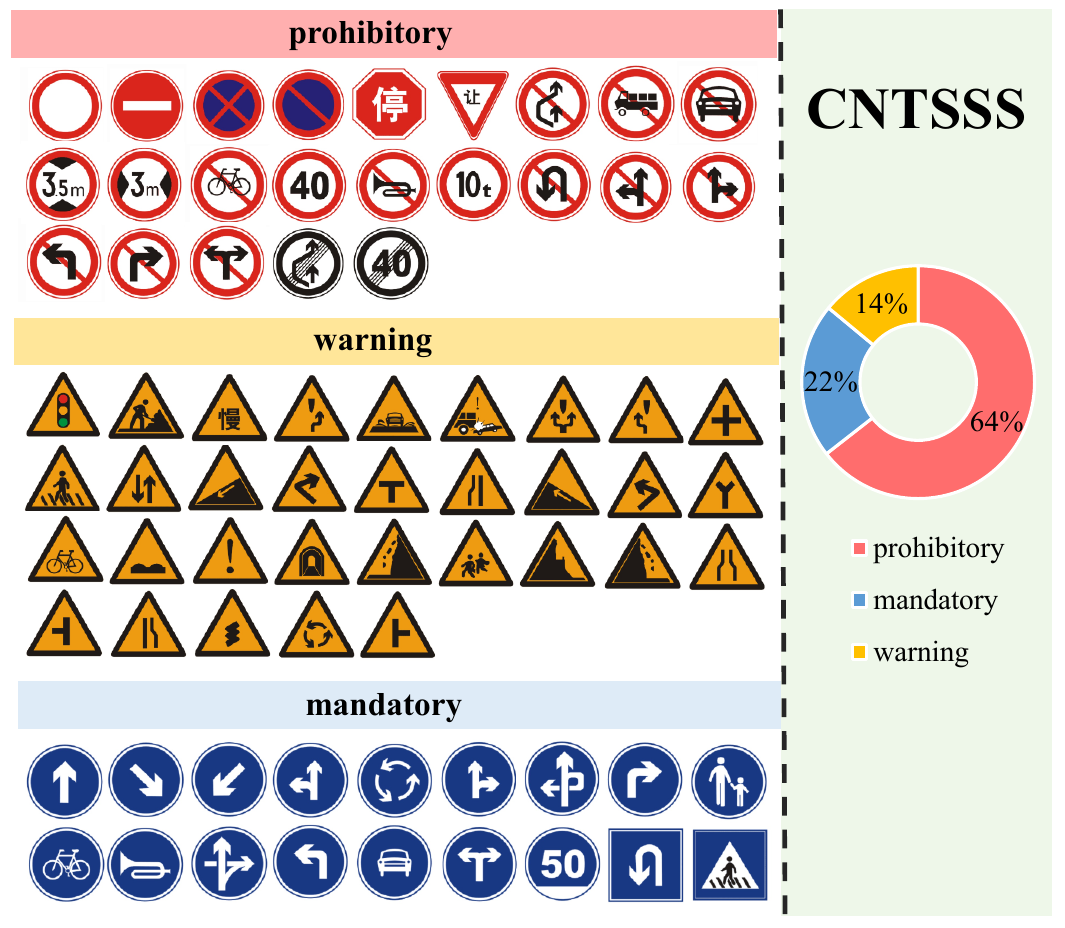} % 调整宽度为栏宽的一半左右
    \caption{The CNTSSS dataset includes three categories of traffic signs: prohibitory signs, mandatory signs, and warning signs. Specifically, prohibitory signs account for 64\% of the collection, mandatory signs account for 22\%, and warning signs account for 14\%.}
    \label{traffic}
\end{figure}

\subsection{Chinese Nighttime Traffic Sign Sample Set}
In the field of traffic sign detection, existing datasets such as TT100K \cite{TT100K-dataset} and GTSRB \cite{GTSRB-dataset} lack samples from nighttime scenes, which limits the performance of detection algorithms under actual nighttime conditions. Although the CCTSDB2021 \cite{CCTSDB-dataset} dataset includes 500 nighttime images for testing, its limited size lacks a large-scale nighttime dataset for training. 

To address this issue, we have collected and constructed a new nighttime traffic sign dataset called Chinese Nighttime Traffic Sign Sample Set (CNTSSS). The dataset includes a total of 4062 nighttime traffic sign images from various cities across China, aiming to provide a rich set of nighttime traffic sign samples to support the research and development of relevant algorithms. As shown in Fig. \ref{city}, the CNTSSS dataset covers scenes from 17 cities across China, including Beijing, Shanghai, Guangzhou, Shenzhen, Jiangmen, Chongqing, Chengdu, Nanchong, Wuhan, Changsha, Tianjin, Nanjing, Zhenjiang, Shangqiu, Shangrao, Guilin, and Jingdezhen. The selection of these cities ensures that the dataset encompasses urban environments from different geographical regions and economic development levels in China, thereby increasing the diversity and representativeness of the data.

The dataset consists of two parts: a training set and a testing set. Considering the  regional diversity, we select four cities—Chengdu, Shanghai, Shenzhen, and Tianjin for testing (786 images), while the remaining 13 cities form the training set (3276 images). 

Following the classification approach of CCTSDB2021, the dataset includes three types of traffic signs as shown in Fig. \ref{traffic}: prohibitory signs (4954 images), mandatory signs (1658 images), and warning signs (1075 images). These three types of signs are the most common on Chinese roadways and are critical for nighttime driving safety. The CNTSSS dataset includes various lighting conditions from dusk to late night, and includes both clear and rainy nighttime weather. Additionally, the dataset covers a range of road scenes, including highways, urban roads, and rural roads. This diversity ensures that researchers can test and optimize their models in different driving environments, improving the practicality and adaptability of their models.

\begin{figure}[t]
    \centering
    \includegraphics[width=0.47\textwidth]{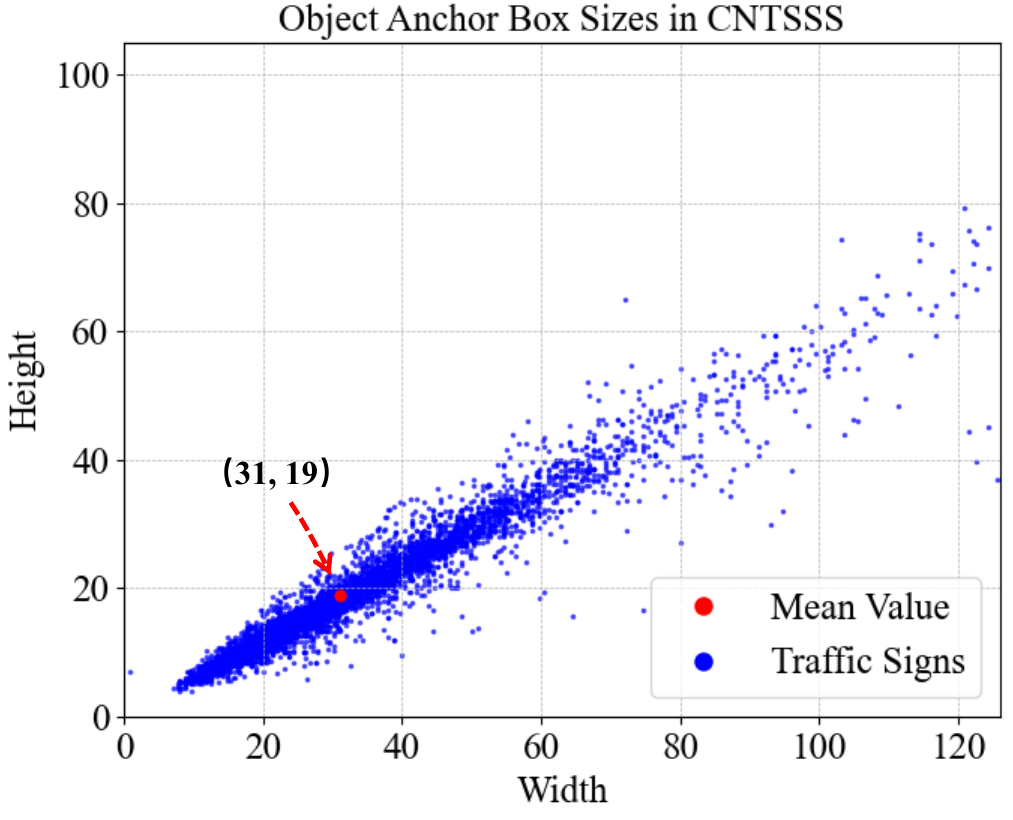} % 调整宽度为栏宽的一半左右
    \caption{The distribution of object anchor box sizes in CNTSSS dataset.}
    \label{count}
\end{figure}

\begin{figure}[t]
    \centering
    \includegraphics[width=0.5\textwidth]{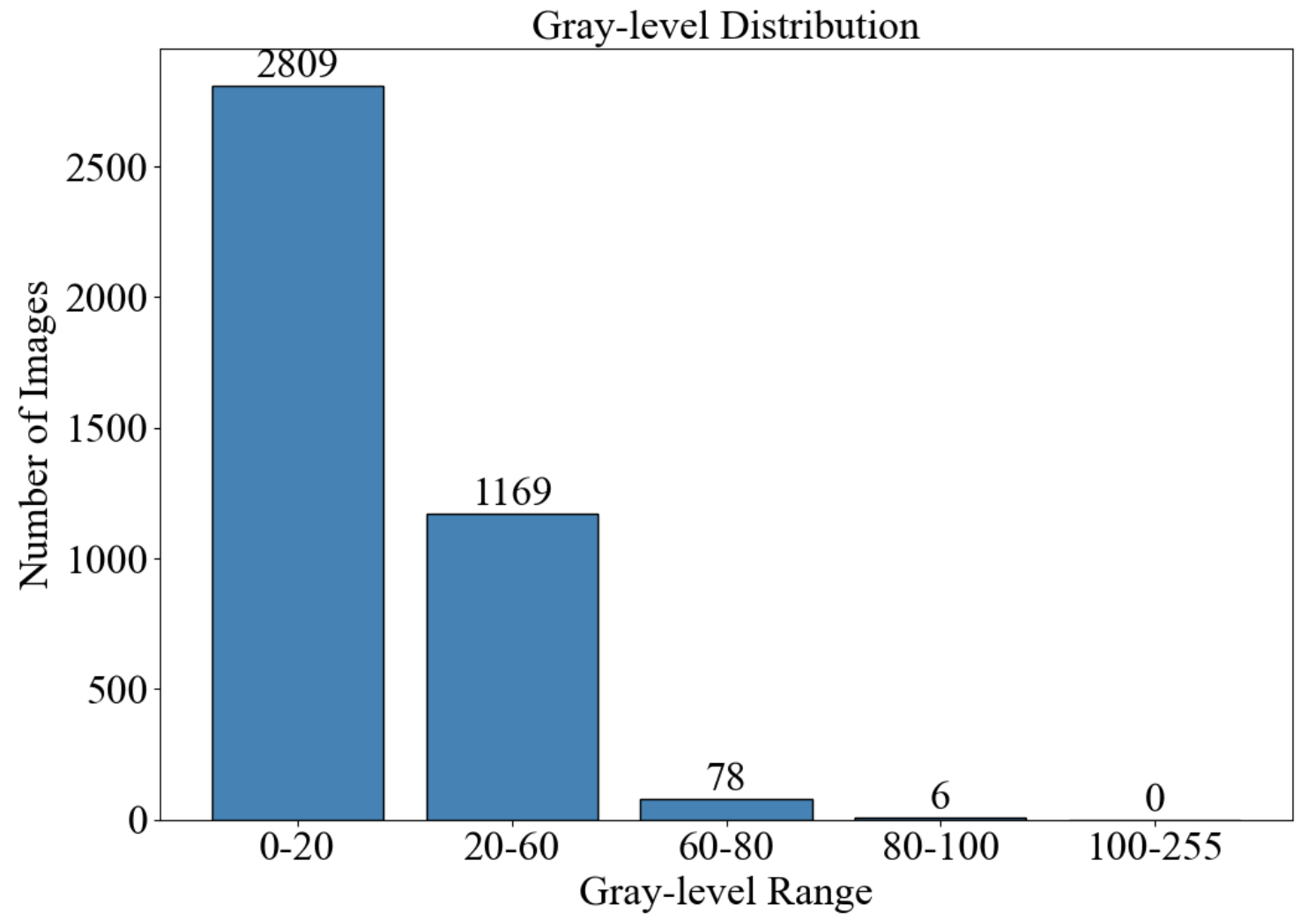} % 调整宽度为栏宽的一半左右
    \caption{The number of low-light images in the CNTSSS dataset is statistically analyzed across different grayscale intervals, where 0 represents black, 255 represents white, and the intermediate values indicate various shades of gray.}
    \label{gray}
\end{figure}

Furthermore, in order to explore the characteristics of the CNTSSS dataset, we conducted an investigation into the object anchor box sizes within the dataset. As shown in Fig. \ref{count}, each blue dot represents the width and height of a traffic sign in pixels. The red dot marks the mean value of the dataset, with coordinates (31, 19), indicating the central tendency of the sign sizes. It can be observed that in our dataset, the traffic signs occupy a relatively small number of pixels. To quantify the lighting differences in nighttime scenes, we calculated the grayscale values of all images in the CNTSSS and divided them into intervals. Grayscale values represent the brightness intensity of each pixel in an image. In grayscale images, each pixel's value is typically a number ranging from 0 to 255, where 0 represents black, 255 represents white, and the intermediate values represent various shades of gray. We converted the original 3-channel BGR images to single-channel grayscale images, then computed the average grayscale value for each image. Grayscale values below 100 can be considered indicative of nighttime images. As shown in Fig. \ref{gray}, images with low illumination dominate the dataset, with no images corresponding to daylight. Experiments demonstrate that our dataset presents a significant challenge for artificial intelligence models.

\begin{figure*}[t]
    \centering
    \includegraphics[width=\textwidth]{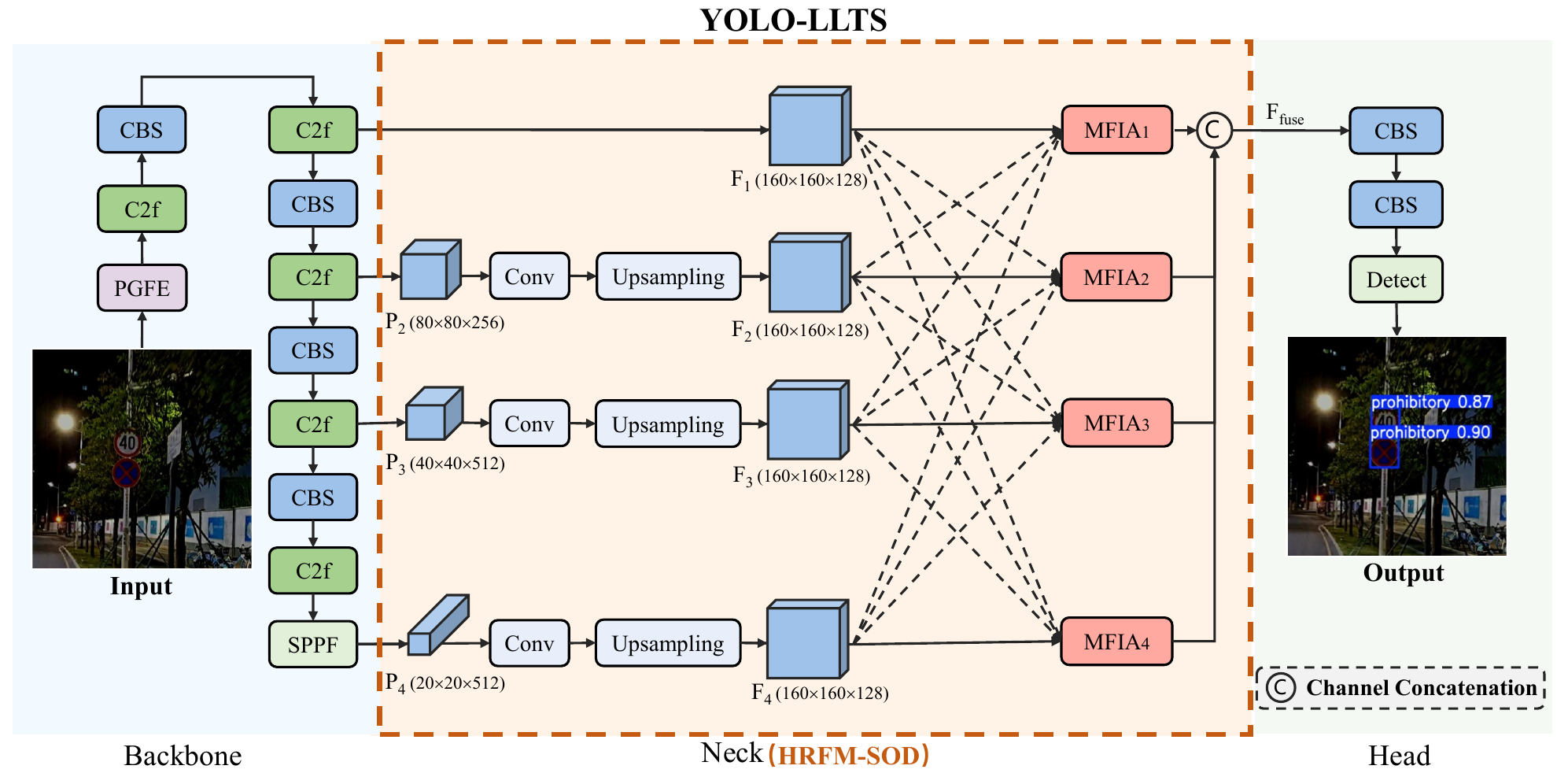}
    \caption{\textbf{Framework overview of our model YOLO-LLTS.} The structure of our YOLO-LLTS model is divided into three components: Backbone, Neck, and Head. The Backbone is responsible for feature extraction, the Neck handles multi-scale feature fusion, and the Head is responsible for final object detection and classification. The main contributions focus on the PGFE module in the Backbone, the introduction of HRFM-SOD as a new Neck, and the MFIA module within the HRFM-SOD.}
    \label{total}
\end{figure*}

\begin{figure*}[t]
    \centering
    \includegraphics[width=\textwidth]{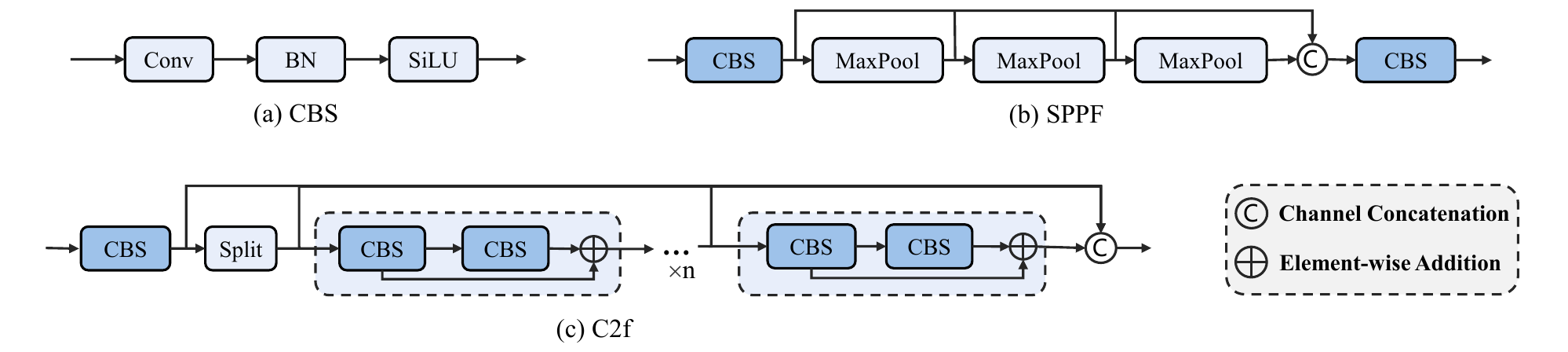}
    \caption{Some module design structures are derived from YOLOv8. (a), (b), and (c) represent the CBS module, SPPF module, and C2f module, respectively.}
    \label{total_in}
\end{figure*}

\subsection{General overview}
Fig. \ref{total} illustrates the overall architecture of the YOLO-LLTS model, which is primarily divided into three components: the backbone network, the neck, and the head.

\textbf{Backbone.} The backbone network is responsible for extracting initial features from the input image. It captures multi-scale important features by utilizing C2f and CBS from YOLOv8. SPPF increases the receptive field, allowing the model to capture more contextual information. The structures of C2f, CBS and SPPF are shown in Fig. \ref{total_in}. The first layer, the PGFE module, represents a key contribution of this work. It is integrated into the backbone network to enhance the quality of low-light images, thus improving the subsequent feature extraction process.

\textbf{Neck.} The neck employs the HRFM-SOD module, which we developed, responsible for multi-scale feature fusion and interaction. It receives the features extracted by the backbone network and refines these features through a series of convolutions and upsampling operations. This module is crucial for enhancing the detection of small objects. The Multibranch Feature Interaction Attention (MFIA) module is embedded within HRFM-SOD to facilitate deep interaction and fusion of multi-scale features.

\textbf{Head.} The head follows the design of YOLOv8 and is responsible for the final object detection and classification. It integrates the refined features from the neck and applies them to predict the position and category of traffic signs within the input image.

The main contributions focus on the PGFE module in the Backbone, the introduction of HRFM-SOD as a new Neck, and the MFIA module within HRFM-SOD. These will be detailed in the subsequent sections, respectively, with a comprehensive discussion of HRFM-SOD, MFIA, and PGFE.

\begin{figure*}[tb]
    \centering
    \includegraphics[width=\textwidth]{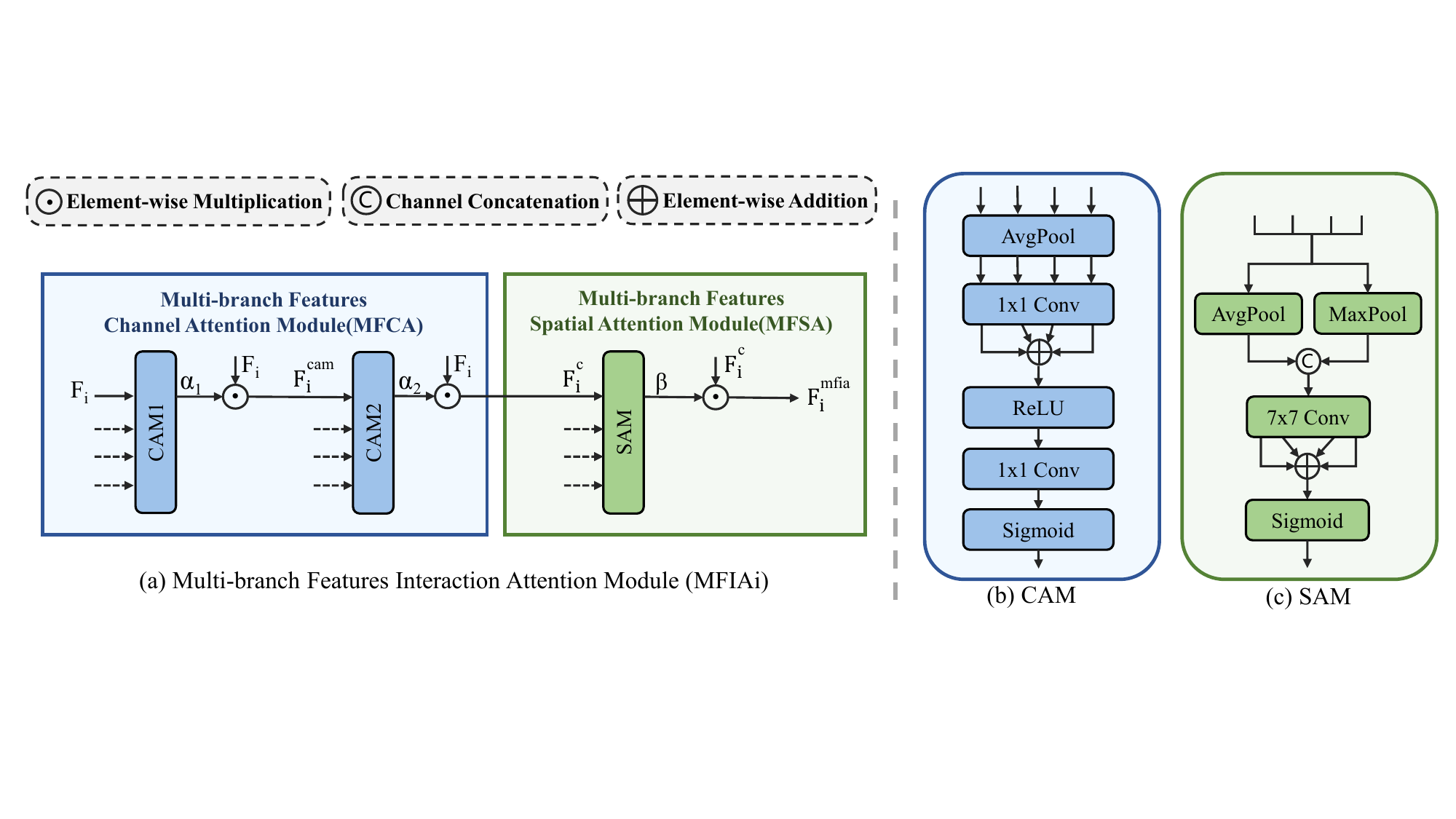}
    \caption{\textbf{Framework overview of Multibranch Features Interaction Attention Module (MFIA).} (a) shows the overall architecture of the MFIA$_i$ module. MFIA is applied to $F_i$($i=1,2,3,4$) four times throughout the network. The illustration shows one MFIA$_i$ module along with its corresponding output feature $F_{i}^{\text{mfia}}$. The MFIA is divided into two parts: the MFCA and MFSA. (b) shows the CAM, which is used twice within the MFCA. (c) illustrates the SAM in the MFSA.}
    \label{MFIA}
\end{figure*}

\subsection{High-Resolution Feature Map for Small Object Detection}

In small object detection under low-light conditions, due to the limited number of pixels occupied by small objects, there is insufficient visual information to highlight their feature representation. The traditional PANet \cite{PANet} has limitations in small object detection, as its feature fusion process may cause the small object's feature information to be diluted or overwhelmed by a large amount of other features, thus affecting the accuracy of small object detection and localization, especially in traffic scenes with dense small objects and complex backgrounds. Furthermore, the fusion of features through top-down and bottom-up methods, involving multiple upsampling and downsampling operations, can lead to information loss.

To address this issue, We introduce the HRFM-SOD module to preserve detailed information in images while simultaneously reducing the network's computational load. As shown in Fig. \ref{total}, deep multi-scale features are uniformly adjusted to the high-resolution image size 160$\times160\times$128 through 1$\times$1 convolution and Bilinear upsampling. The formula is as follows:
\begin{align}
F_2 &= \operatorname{Upsampling}\left(\operatorname{Conv}\left(P_2\right)\right) \label{eq:F2} \\
F_3 &= \operatorname{Upsampling}\left(\operatorname{Conv}\left(P_3\right)\right) \label{eq:F3} \\
F_4 &= \operatorname{Upsampling}\left(\operatorname{Conv}\left(P_4\right)\right) \label{eq:F4}
\end{align}
where P$_2$, P$_3$ and P$_4$ represent the input features, Conv$(\cdot)$ denotes the convolution operation, and Upsampling$(\cdot)$ refers to the upsampling operation. F$_2$, F$_3$, and F$_4$ represent the output features.

In the feature extraction process, we adjust the receptive field obtained from the backbone using the computational method proposed in \cite{25}. As a result, the receptive field received by the HRFM-SOD module varies with the size of the target.

The high-resolution feature maps of the same scale are fed into the MFIA module (MFIA$_1$ to MFIA$_4$) for feature fusion. This module is designed to focus more on the resolution of small objects. By integrating the feature maps {$F_{1}$, $F_{2}$, $F_{3}$, $F_{4}$} from four different receptive fields, it significantly enhances the feature representation of small objects while reducing computational costs. Finally, the four features are fused through concatenation, as shown in the following formula:
\begin{equation}
F_{\operatorname{fuse}}=\operatorname{Concat}\left(\operatorname{MFIA}_i\left(F_1, F_2, F_3, F_4\right)\right)
\end{equation}
where $i$ represents the $i$-th pass through the MFIA ($i=1,2,3,4$).

\begin{figure*}[htb]
    \centering
    \includegraphics[width=\textwidth]{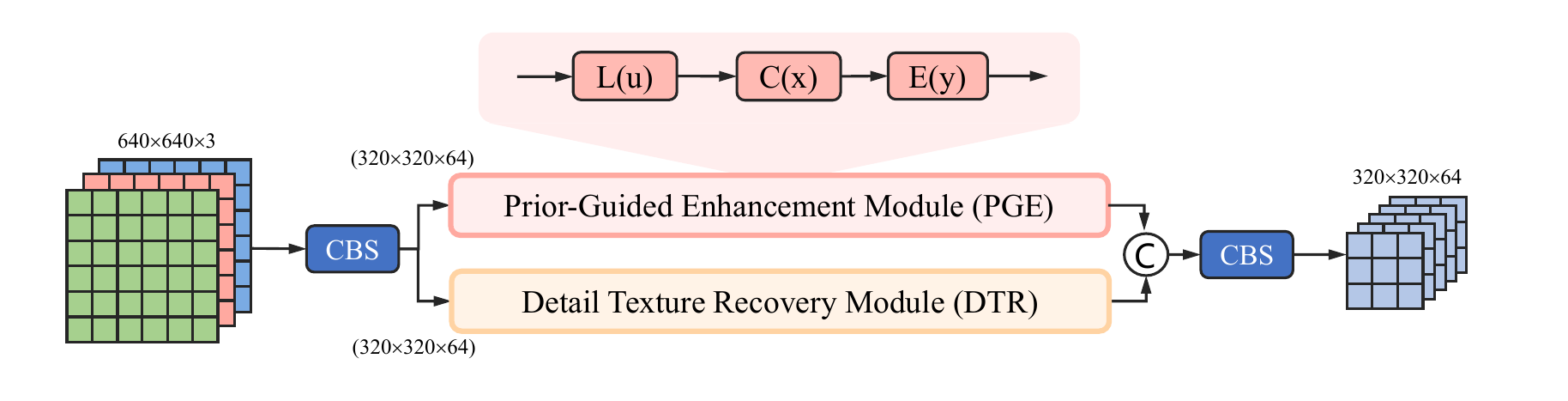}
    \caption{\textbf{Overview of Prior-Guided Feature Enhancement Module (PGFE).} The PGFE module is divided into two parts: the Prior-Guided Enhancement Module (PGE) and the Detail Texture Recovery Module (DTR).}
    \label{PGFE}
\end{figure*}

\subsection{Multibranch Features Interaction Attention Module}
In low-light conditions, small targets have low resolution and contain less information, requiring the model to have a stronger ability to capture information. The attention mechanism has been extensively studied and applied to enhance the model's ability to capture key features (e.g., SENet \cite{SENet}, CANet \cite{CANet}, CBAM \cite{CBAM}). However, most existing attention mechanisms focus on processing a single feature, neglecting the potential complementarity and interactions between features. Dai et al. \cite{AFF} proposed a method for attentional feature fusion of local and global features, yet this method failed to address the challenge of fusing features across more than two scales. Zhao et al. \cite{BA-Net} proposed BA-Net, which improved SENet by leveraging information from shallow convolutional layers but overlooked the spatial domain of the image. Therefore, we propose an attention mechanism called Multibranch Features Interaction Attention Module (MFIA). 

As shown in Fig. \ref{MFIA}, MFIA consists of two components: the Multibranch Features Channel Attention Module (MFCA) and the Multibranch Features Spatial Attention Module (MFSA). It effectively utilizes features with semantic and scale inconsistencies, enabling efficient interaction between multiple features in both the channel and spatial dimensions. Notably, MFIA is employed four times throughout the network. The illustration shows one MFIA$_i$ module along with its corresponding output feature $F_{i}^{\text{mfia}}$.
\subsubsection{Multibranch Features Channel Attention Module}
MFCA$_i$ achieves preliminary interaction of multi-receptive field features through two consecutive Channel Attention Modules (CAM$_1$ and CAM$_2$). Each CAM layer processes the input features using 1$\times$1 convolutions followed by ReLU activation functions, and then generates attention weights $\alpha_{1}$ and $\alpha_{2}$ sequentially through the Sigmoid function. These weights are used to adjust the channel importance of the input features. Our experiments demonstrate that by iterating through lightweight channel attention layers multiple times, feature interactions can be enhanced without compromising model performance, thus avoiding the potential bottleneck that may arise from the initial feature map integration. Additionally, both weight multiplications are applied to the initial feature $F_{i}$ to prevent the loss of original information during feature transmission. The entire process can be expressed as:
\begin{equation}
\left\{
\begin{array}{l}
\alpha_n = \operatorname{CAM}_n(F_1, F_2, F_3, F_4) \\
F_i^{\operatorname{cam}} = \alpha_n \cdot F_i
\end{array}
\right.
\end{equation}
where, $n$ denotes the $n$-th pass through the CAM ($n=1, 2$), and $i$ represents the $i$-th pass through the MFIA ($i=1,2,3,4$).
\subsubsection{Multibranch Features Spatial Attention Module}
The MFSA module achieves deep interaction between feature maps through the Spatial Attention Mechanism (SAM). SAM extracts spatial information from the feature maps using average pooling and max pooling layers, followed by further processing of this information through a 7$\times$7 convolutional layer. Finally, the spatial attention weight $\beta$ is generated through a Sigmoid function. This weight is used to adjust the spatial importance of the feature maps, and the process can be expressed as:
\begin{equation}
\left\{
\begin{array}{l}
\beta=\operatorname{SAM}(F_1^{c}, F_2^{c}, F_3^{c}, F_4^{c}) \\
F_i ^{\operatorname{mfia}}=\beta \cdot F_i^{c}
\end{array}
\right.
\end{equation}
where $i$ denotes the $i$-th pass through the MFIA ($i=1,2,3,4$).

The MFIA module enhances the model's ability to capture important features by facilitating interactions between features of different scales (i.e., varying sizes or resolutions). It is similar to how the human eye adjusts its focus based on the distance and importance of objects when observing a scene. In this way, the model is able to focus on critical details, especially in complex images such as small and blurry traffic signs, while reducing the interference from irrelevant parts.

\begin{figure}[t]
    \centering
    \includegraphics[width=0.5\textwidth]{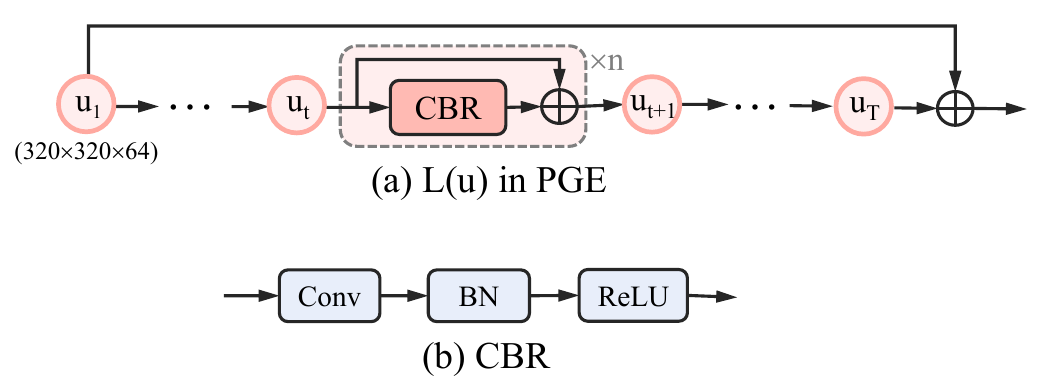}
    \caption{(a) shows the structure of L(u) in the PGE, which enhances features while improving image brightness. The pink dashed box represents G(·) from Eq. \ref{eq:G}, denoting the residual operation. (b) shows the structure of the CBR, which is used multiple times within the residual operation in L(u).}
    \label{PGE}
\end{figure}

\begin{figure*}[t]
    \centering
    \includegraphics[width=\textwidth]{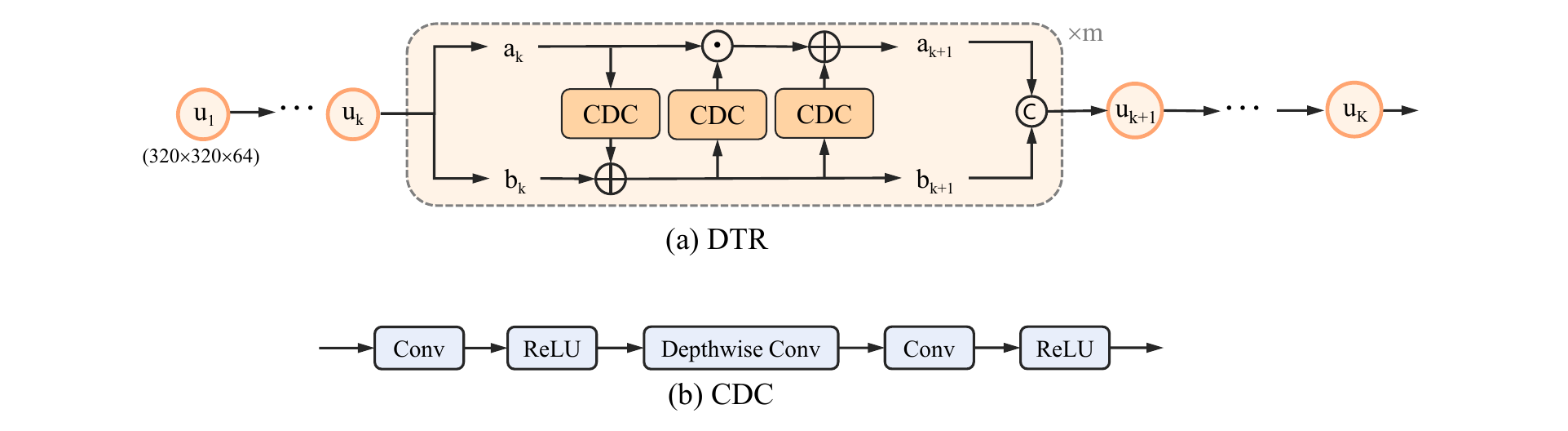}
    \caption{(a) shows the structure of the DTR in the PGFE, which is employed to supplement the detailed information lost due to feature enhancement. (b) shows the structure of the CDC in the DTR, where the CDC utilizes Depthwise Convolution to extract local features from the image.}
    \label{DTR}
\end{figure*}

\subsection{Prior-Guided Feature Enhancement Module}
Images captured in conditions of diminished lighting often suffer from poor quality, which manifests as increased noise, reduced contrast, blurred edges, and hidden information in dark areas, significantly affecting the accuracy of traffic sign detection. However, simply using existing networks that increase exposure may amplify image noise, leaving the originally low-quality image still unclear, which negatively impacts subsequent object detection tasks. 

To address this challenge, we propose the Prior-Guided Feature Enhancement (PGFE) Module. This module transforms RGB images from 3 channels to 64 channels to enhance low-light images and provide prior knowledge for the following detection. As shown in Fig. \ref{PGFE}, the module consists of two main components: the Prior-Guided Enhancement (PGE) Module and the Detail Texture Recovery (DTR) Module.

\subsubsection{Prior-Guided Enhancement Module} Retinex theory \cite{Retinex} indicates that there is a connection between the desired clear image $z$ and low-light observation $y$: $y=z\otimes x$, where $x$ represents the illumination component. Inspired by SCI \cite{SCI}, the image brightness can be improved to some extent by learning a residual representation between illumination and low-light observation. Compared to directly mapping low-light observation to illumination, learning the residual representation significantly reduces computational complexity and helps avoid exposure issues caused by imbalanced brightness enhancement. This design allows our network to maintain strong performance at night while preserving its performance during the day. As shown in Fig. \ref{PGE}, $L(u)$ applies n consecutive residual operations on the input features, and then the final output is added to the initial features to enhance them. The formula is as follows:
\begin{equation}
\left\{\begin{array}{l}
G\left(u_t\right)=u_{t-1}+\operatorname{CBR}\left(u_{t-1}\right) \\
L(u)=u_1+G\left(u_t\right)
\end{array}\right.
\label{eq:G}
\end{equation}
where $u_{t}$ represents the output parameters at the $t$-th stage ($t=0,…,T$), and $G(\cdot)$ denotes the residual operation. $u_{1}$ represents the input feature, while $u_{T}$ represents the output feature.

The output of $L(\cdot)$ is fed into $C(\cdot)$ for contrast enhancement, and then into $E(\cdot)$for edge enhancement. The principles of both operations are similar. $C(\cdot)$ enhances the contrast by amplifying the differences between different pixel values, and this can be represented by the following formula:
\begin{equation}
C(x)=\gamma \cdot(x-\bar{x})+\bar{x}
\end{equation}
where $\overline{x}$ represents the mean value of $x$, and $\gamma$ is the contrast enhancement coefficient, which is set to 2.
$E(y)$ enhances the edges by amplifying the difference between the input image and its blurred version obtained through Gaussian filtering. The formula can be expressed as:
\begin{equation}
E(y)=\delta \cdot|y-\operatorname{Gaussian}(y)|+y
\end{equation}
where $Gaussian(\cdot)$ denotes the Gaussian filter, and $\delta$ represents the sharpening intensity coefficient, which is set to 2.5.

\subsubsection{Detail Texture Recovery Module} Feature enhancement may lead to the loss of original information in low-light images, thus requiring a network capable of effectively extracting image details and textures to supplement this. Inspired by \cite{High-frequency, INN, INN2, cddfuse}, the Invertible Neural Network (INN) extracts local information that is highly correlated with the high-frequency features in the frequency domain, specifically the edges and lines within the image. INN prevents information loss by dividing the input parameters into two parts, allowing the input and output features to mutually generate each other. This can be considered as a lossless feature extraction, which is particularly suitable in this context. As shown in Fig. \ref{DTR}, our DTR iterates the INN m times to achieve complete extraction of detail textures. Specifically, by splitting the input feature $u_{k}$ into two parts, $a_{k}$ and $b_{k}$, the process becomes reversible, ensuring lossless transmission of detail features while simultaneously supplementing the enhanced feature map. Therefore, $a_{k+1}$ and $b_{k+1}$ can be expressed in terms of $a_{k}$ and $b_{k}$, as shown in the following formula:
\begin{equation}
\left\{\begin{array}{l}
b_{k+1} = b_k+\operatorname{CDC}\left(a_k\right) \\
a_{k+1} = a_k \cdot \exp \left(\operatorname{CDC}\left(b_{k+1}\right)\right)+\operatorname{CDC}\left(b_{k+1}\right)
\end{array}\right.
\label{eq:G}
\end{equation}

The next stage $u_{k+1}$ of $u_{k}$ is obtained by concatenating $a_{k+1}$ and $b_{k+1}$:

\begin{equation}
\begin{aligned}
u_{k+1}&=\operatorname{Concat}\left(a_{k+1}, b_{k+1}\right)
\end{aligned}
\end{equation}

where $u_{k}$ represents the output parameters at the $k$-th stage ($k=0,…,T$). As shown in Fig. \ref{DTR}, $u_{1}$ represents the input features, while $u_{K}$ represents the output features.

The DTR module supplements the features extracted from the original image with detailed features to enhance the features processed by the PGE module. This design allows the PGFE module to enhance the image without losing any details.

\begin{figure}[t]
    \centering
    \includegraphics[width=0.48\textwidth]{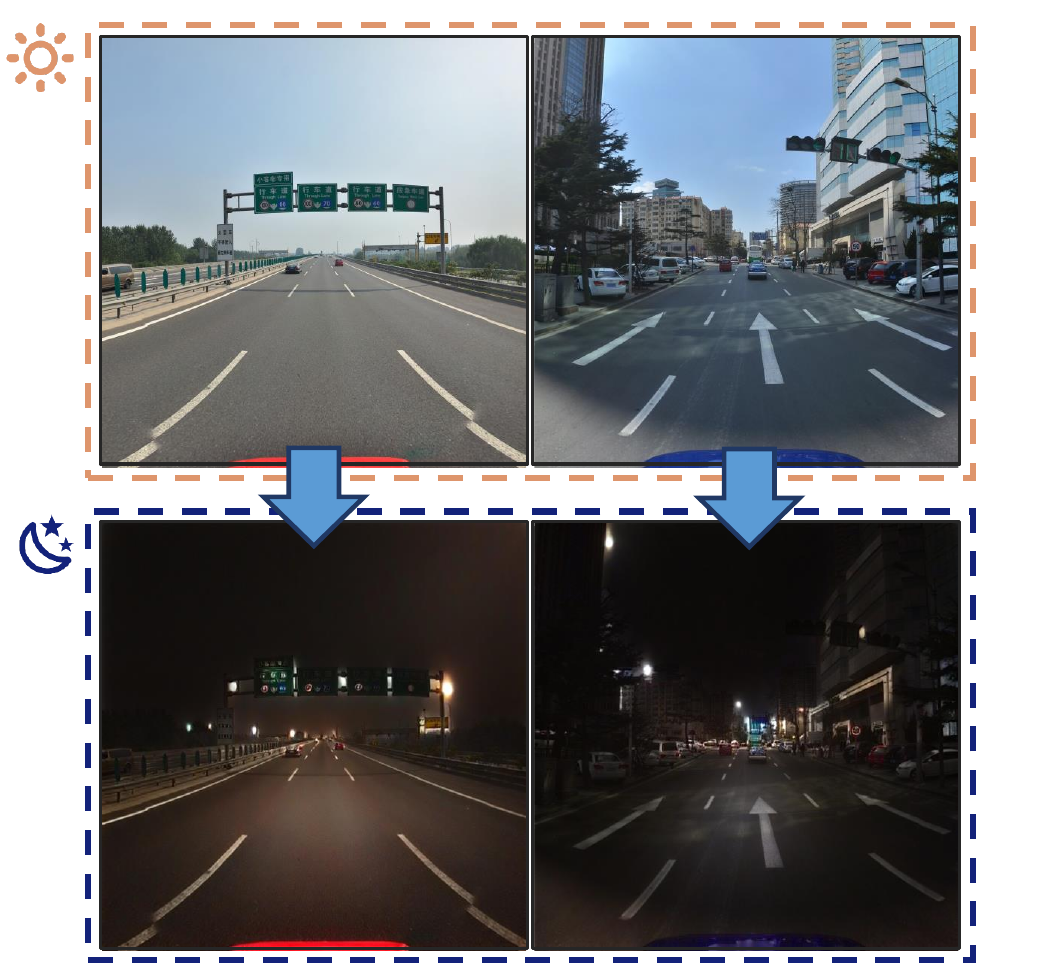} % 调整宽度为栏宽的一半左右
    \caption{Use CycleGAN to generate a low-light image dataset, TT100K-night, from the TT100K dataset.}
    \label{tt100k-night}
\end{figure}

\section{Experiments}
\label{sec:experiments}
In this section, we provide a detailed description of the dataset, parameter settings, and evaluation metrics used in the experiments. The effectiveness of the algorithm and the rationality of the structure are demonstrated through experimental results. Finally, we perform error analysis on the experimental outcomes and conduct real-world testing. 
\subsection{Datasets}
To evaluate the performance of our model in recognizing traffic signs under nighttime conditions, we conduct a comprehensive assessment using the publicly available datasets TT100K, CCTSDB2021, and our proprietary dataset CNTSSS.
\subsubsection{TT100K-night}The TT100K dataset, organized and released by the joint laboratory of Tsinghua University and Tencent, is modified following the approach of Zhu et al. \cite{TT100K-dataset}, where categories with fewer than 100 samples were excluded, narrowing the focus to 45 categories. The training set consists of 6,105 images, while the test set includes 3,071 images, each with a resolution of $2048\times2048$ pixels. As shown in Fig.\ref{tt100k-night}, we employed CycleGAN \cite{cycleGAN} to augment the TT100K dataset, ensuring a more accurate evaluation of the model's performance.
\subsubsection{CNTSSS}Converting daytime data to nighttime conditions cannot accurately assess a model's performance during the night. Therefore, we have constructed our own dataset consisting of traffic sign images captured exclusively at night. The CNTSSS dataset includes 3,276 images in the training set and 786 images in the test set, with traffic signs classified into mandatory, prohibitory, or warning types. Further details have been outlined in the previous section.
\subsubsection{CCTSDB2021}The CCTSDB2021 dataset was created by Changsha University of Science and Technology and comprises 17,856 images in both the training and test sets, with traffic signs classified into mandatory, prohibitory, or warning types. The training set contains 16,356 images, of which approximately 700 are captured at night, while the remaining images are taken during the day. Although the training set does not include as much nighttime data as the CNTSSS dataset, this distribution better reflects real-world driving conditions, making it a challenging yet valuable benchmark. To assess the model's performance at night, we selected 500 nighttime images from the test set as the basis for performance evaluation.
\subsubsection{GTSDB-night}The GTSDB dataset consists of 900 images, with 600 images allocated for training and 300 images for testing. This dataset covers 43 common traffic sign categories found in Germany. Each image has a resolution of $1360\times800$ pixels and may contain one or more traffic signs, or none at all. Similarly, we use CycleGAN to generate nighttime images from daytime images in order to evaluate the model's performance under low-light conditions.

\begin{table*}[tb]
\caption{Performance comparison on \textbf{TT100K-night} dataset; The first and second best results are indicated in \textbf{\textcolor{blue}{blue}} and \textbf{\textcolor{green}{green}}.}
\label{tt100k}
\renewcommand{\arraystretch}{1.5}
\begin{adjustbox}{width=\textwidth}
\begin{tabular}{cccccccccc>{\columncolor[HTML]{EFEFEF}}c}
\Xhline{1pt}
        Method & Venue & Input Size & Precision(\%) & Recall(\%) & F1 & mAP50(\%) & mAP50:95(\%) & Param(M)$\downarrow$ & FPS$\uparrow$ & GPU \\
\hline
        YOLOv5-L\cite{yolov5} & 2020 & 640$\times$640 & 70.6 & 57.2 & 63.2 & 62.4 & 44.4 & 46.1 & 99.0 & RTX 4090 \\
        YOLOv6-L\cite{yolov6} & CVPR2022 & 640$\times$640 & 65.7 & 52.8 & 58.5 & 57.4 & 40.9 & 59.5 & 43.0 & RTX 4090 \\
        GOLD-YOLO-L\cite{gold-yolo} & NeurIPS2023 & 640$\times$640 & 71.3 & 54.7 & 61.9 & 60.5 & 43.1 & 75.0 & 55.2 & RTX 4090 \\
        YOLOv8-L\cite{yolov8_ultralytics} & 2023 & 640$\times$640 & 72.1 & 59.3 & 65.1 & 65.2 & 46.3 & 43.6 & \textbf{\textcolor{green}{107.5}} & RTX 4090 \\
        YOLOv9-C\cite{yolov9} & CVPR2024 & 640$\times$640 & 71.3 & 59.7 & 65.0 & 64.6 & 45.5 & 50.7 & 80.6 & RTX 4090 \\
        YOLOv10-L\cite{yolov10} & NeurIPS2024 & 640$\times$640 & 68.4 & 57.4 & 62.4 & 61.4 & 43.8 & 25.7 & 72.5 & RTX 4090 \\
        Zhang et al. \cite{Zhang2024} & TETCI2024 & 640$\times$640 & 17.3 & 18.9 & 18.1 & 21.9 & - & 20.8 & 105.1 & RTX 4090 \\
        MIAF-net \cite{23} & TIM2024 & 640$\times$640 & 69.6 & 57.7 & 63.1 & 60.4 & 41.2 & 34.4 & 102.0 & RTX 4090 \\
        YOLOv11-L\cite{yolo11_ultralytics} & 2024 & 640$\times$640 & 69.5 & 57.3 & 62.8 & 62.5 & 44.1 & 25.3 & 73.0 & RTX 4090 \\
        YOLO-TS\cite{25} & arXiv2024 & 640$\times$640 & \textbf{\textcolor{green}{75.2}} & \textbf{\textcolor{green}{63.2}} & \textbf{\textcolor{green}{68.7}} & \textbf{\textcolor{green}{68.7}} & \textbf{\textcolor{green}{48.4}} & \textbf{\textcolor{green}{11.1}} & \textbf{\textcolor{blue}{109.9}} & RTX 4090 \\
        YOLOv12-L\cite{yolov12} & arXiv2025 & 640$\times$640 & 72.6 & 54.5 & 62.3 & 61.4 & 43.8 & 26.4 & 46.3 & RTX 4090 \\
\hline
        YOLO-LLTS(ours) & - & 640$\times$640 & \textbf{\textcolor{blue}{77.2(+2.0)$\uparrow$}} & \textbf{\textcolor{blue}{64.4(+1.2)$\uparrow$}} & \textbf{\textcolor{blue}{70.2(+1.5)$\uparrow$}} & \textbf{\textcolor{blue}{71.4(+2.7)$\uparrow$}} & \textbf{\textcolor{blue}{50.0(+1.6)$\uparrow$}} & \textbf{\textcolor{blue}{9.9}} & 83.3 & RTX 4090 \\
\Xhline{1pt}
\end{tabular}
\end{adjustbox}

\begin{tablenotes}
    \footnotesize
    \item Note: The \textbf{\textcolor{blue}{bold blue}} represents the optimal result, \textbf{\textcolor{green}{blod green}} represents the suboptimal value, and differences less than or equal to 0.4\% are marked with a \text{\textdagger} symbol.
\end{tablenotes}
\end{table*}

\subsection{Experimental Settings}
\subsubsection{Training details}Our experiments are carried out on a machine with four NVIDIA GeForce RTX 4090 GPUs. The input images are resized to a resolution of $640\times640$ pixels. The number of epochs for training is set to 200 for the CNTSSS dataset and 300 for the TT100K and CCTSDB2021 datasets. The batch size is set to 48. We utilize Stochastic Gradient Descent (SGD) with a learning rate of 0.01 and a momentum of 0.937.
\subsubsection{Evaluation metrics}We evaluate the performance of the proposed algorithm using Precision, Recall, F1 score, mean Average Precision at 50\% IoU (mAP50), mean Average Precision from 50\% to 95\%  IoU (mAP50:95), and speed (FPS). These metrics \cite{quantification} are calculated using the following formulas: 
\begin{align}
\text { Precision } &=\frac{\mathrm{TP}}{\mathrm{TP}+\mathrm{FP}} 
\end{align}
\begin{align}
\text { Recall } &=\frac{\mathrm{TP}}{\mathrm{TP}+\mathrm{FN}}
\end{align}
\begin{align}
\text { F1 } &=\frac{2 \times \text { Precision } \times \text { Recall }}{\text { Precision }+ \text { Recall }}
\end{align}
\begin{align}
\text{mAP50} &=\frac{1}{N} \sum_{i=1}^N \mathrm{AP}_{50}^i
\end{align}
\begin{align}
\text{mAP50:95} &=\frac{1}{M} \sum_{k=1}^M\left(\frac{1}{N} \sum_{i=1}^N \mathrm{P}_{50: 95}^{i, k}\right)
\end{align}

where TP represents the number of correctly identified positive instances. FP refers to the count of instances incorrectly identified as positive. FN represents the count of instances incorrectly identified as negative. $N$ represents the total number of categories. $M$ represents the number of Intersection over Union (IoU) threshold intervals which is equal to 10, spanning from 0.5 to 0.95 with an increment of 0.05.

\subsection{Comparisons with the State-of-the-arts}
In comparison with existing state-of-the-art technologies, our model demonstrates superior performance in both accuracy and speed. We compare YOLO-LLTS with several models, including YOLOv5 \cite{yolov5}, YOLOv6 \cite{yolov6}, GOLD-YOLO \cite{gold-yolo}, YOLOv8 \cite{yolov8_ultralytics}, YOLOv9 \cite{yolov9}, YOLOv10 \cite{yolov10}, Zhang et al. \cite{Zhang2024}, MIAF-net \cite{23}, YOLOv11 \cite{yolo11_ultralytics}, YOLO-TS \cite{25}, and YOLOv12 \cite{yolov12}. These comparison models are all advanced models developed in the past five years.

\begin{table*}[htb]
\caption{Performance comparison on \textbf{CNTSSS} dataset; The first and second best results are indicated in \textbf{\textcolor{blue}{blue}} and \textbf{\textcolor{green}{green}}.}
\label{cntsss}
\renewcommand{\arraystretch}{1.5}
\begin{adjustbox}{width=\textwidth}
\begin{tabular}{cccccccccc>{\columncolor[HTML]{EFEFEF}}c}
\Xhline{1pt}
        Method & Venue & Input Size & Precision(\%) & Recall(\%) & F1 & mAP50(\%) & mAP50:95(\%) & Param(M)$\downarrow$ & FPS$\uparrow$ & GPU \\
\hline
        YOLOv5-L\cite{yolov5} & 2020 & 640$\times$640 & 83.7 & 66.7 & 74.2 & 75.0 & 53.4 & 46.1 & 97.1 & RTX 4090 \\
        YOLOv6-L\cite{yolov6} & CVPR2022 & 640$\times$640 & 86.4 & 73.1 & 79.2 & 79.8 & 55.9 & 59.5 & 45.2 & RTX 4090 \\
        GOLD-YOLO-L\cite{gold-yolo} & NeurIPS2023 & 640$\times$640 & 86.0 & \textbf{\textcolor{green}{73.9}} & \textbf{\textcolor{green}{79.5}} & \textbf{\textcolor{green}{79.9}} & 56.2 & 75.0 & 42.9 & RTX 4090 \\
        YOLOv8-L\cite{yolov8_ultralytics} & 2023 & 640$\times$640 & 85.1 & 67.7 & 75.4 & 75.1 & 53.3 & 43.6 & 100.0 & RTX 4090 \\
        YOLOv9-C\cite{yolov9} & CVPR2024 & 640$\times$640 & 84.1 & 66.2 & 74.1 & 74.7 & 52.8 & 50.7 & 80.0 & RTX 4090 \\
        YOLOv10-L\cite{yolov10} & NeurIPS2024 & 640$\times$640 & 83.9 & 63.8 & 72.5 & 72.6 & 52.1 & 25.7 & 78.1 & RTX 4090 \\
        Zhang et al. \cite{Zhang2024} & TETCI2024 & 640$\times$640 & 42.7 & 44.6 & 43.6 & 48.9 & - & 20.8 & 110.8 & RTX 4090 \\
        MIAF-net \cite{23} & TIM2024 & 640$\times$640 & 84.1 & 69.0 & 75.8 & 76.2 & 52.6 & 34.2 & \textbf{\textcolor{blue}{158.7}} & RTX 4090 \\
        YOLOv11-L\cite{yolo11_ultralytics} & 2024 & 640$\times$640 & 80.4 & 66.9 & 73.0 & 73.2 & 51.9 & 25.3 & 65.4 & RTX 4090 \\
        YOLO-TS\cite{25} & arXiv2024 & 640$\times$640 & \textbf{\textcolor{green}{87.3}} & 70.9 & 78.2 & 79.6 & \textbf{\textcolor{green}{58.2}} & \textbf{\textcolor{green}{15.1}} & \textbf{\textcolor{green}{117.6}} & RTX 4090 \\
        YOLOv12-L\cite{yolov12} & arXiv2025 & 640$\times$640 & 85.0 & 62.6 & 72.1 & 72.8 & 51.7 & 26.3 & 46.1 & RTX 4090 \\
\hline
        YOLO-LLTS(ours) & - & 640$\times$640 & \textbf{\textcolor{blue}{88.3(+1.0)$\uparrow$}} & \textbf{\textcolor{blue}{74.9(+1.0)$\uparrow$}} & \textbf{\textcolor{blue}{81.0(+1.5)$\uparrow$}} & \textbf{\textcolor{blue}{81.2(+1.3)$\uparrow$}} & \textbf{\textcolor{blue}{60.1(+1.9)$\uparrow$}} & \textbf{\textcolor{blue}{13.9}} & 82.0 & RTX 4090 \\
\Xhline{1pt}
\end{tabular}
\end{adjustbox}
\begin{tablenotes}
    \footnotesize
    \item Note: The article \cite{Zhang2024} only released the trained weights for the CCTSDB2021 dataset. Therefore, experimental results for datasets other than CCTSDB2021 were obtained by training the model using the open-source code.
\end{tablenotes}
\end{table*}

\begin{table*}[htb]
\caption{Performance comparison on \textbf{CCTSDB2021} dataset; The first and second best results are indicated in \textbf{\textcolor{blue}{blue}} and \textbf{\textcolor{green}{green}}.}
\label{cctsdb}
\renewcommand{\arraystretch}{1.5}
\begin{adjustbox}{width=\textwidth}
\begin{tabular}{cccccccccc>{\columncolor[HTML]{EFEFEF}}c}
\Xhline{1pt}
        Method & Venue & Input Size & Precision(\%) & Recall(\%) & F1 & mAP50(\%) & mAP50:95(\%) & Param(M)$\downarrow$ & FPS$\uparrow$ & GPU \\
\hline
        YOLOv5-L\cite{yolov5} & 2020 & 640$\times$640 & 85.4 & 72.2 & 78.2 & 78.6 & 50.2 & 46.1  & 88.5 & RTX 4090 \\
        YOLOv6-L\cite{yolov6} & CVPR2022 & 640$\times$640 & 87.4 & 76.5 & 81.6 & 82.0 & 52.6 & 59.5 & 52.2 & RTX 4090 \\
        GOLD-YOLO-L\cite{gold-yolo} & NeurIPS2023 & 640$\times$640 & 83.8 & 77.0 & 80.3 & 80.6 & 51.2 & 75.0 & 63.3 & RTX 4090 \\
        YOLOv8-L\cite{yolov8_ultralytics} & 2023 & 640$\times$640 & 84.4 & 74.4 & 79.1 & 80.5 & 53.0 & 43.6 & 91.7 & RTX 4090 \\
        YOLOv9-C \cite{yolov9} & CVPR2024 & 640$\times$640 & 84.8 & 76.0 & 80.2 & 81.9 & 53.7 & 50.7 & 42.4 & RTX 4090 \\
        YOLOv10-L\cite{yolov10} & NeurIPS2024 & 640$\times$640 & 83.9 & 73.2 & 78.2 & 78.7 & 51.4 & 25.7 & 68.0 & RTX 4090 \\
        Zhang et al. \cite{Zhang2024} & TETCI2024 & 640$\times$640 & 82.7 & 80.7 & 81.6 & 78.1 & - & 20.8 & 101.1 & RTX 4090 \\
        MIAF-net \cite{23} & TIM2024 & 640$\times$640 & 60.1 & 49.6 & 54.5 & 52.7 & 36.2 & 34.4 & \textbf{\textcolor{blue}{151.5}} & RTX 4090 \\
        YOLOv11-L\cite{yolo11_ultralytics} & 2024 & 640$\times$640 & 85.8 & 74.2 & 79.6 & 81.2 & 54.1 & 25.3 & 70.4 & RTX 4090 \\
        YOLO-TS\cite{25} & arXiv2024 & 640$\times$640 & \textbf{\textcolor{green}{88.1}} & \textbf{\textcolor{green}{80.8}} & \textbf{\textcolor{green}{84.3}} & \textbf{\textcolor{green}{86.0}} & \textbf{\textcolor{green}{57.2}} & \textbf{\textcolor{green}{12.9}} & \textbf{\textcolor{green}{138.9}} & RTX 4090 \\
        YOLOv12-L\cite{yolov12} & arXiv2025 & 640$\times$640 & 84.9 & 74.2 & 79.2 & 82.8 & 55.7 & 26.3 & 44.8 & RTX 4090 \\
\hline
        YOLO-LLTS(ours) & - & 640$\times$640 & \textbf{\textcolor{blue}{88.8(+0.7)}} & \textbf{\textcolor{blue}{81.1$^{\text{\textdagger}}$(+0.3)}} & \textbf{\textcolor{blue}{84.8(+0.5)}} & \textbf{\textcolor{blue}{87.8(+1.8)$\uparrow$}} & \textbf{\textcolor{blue}{57.5$^{\text{\textdagger}}$(+0.3)}} & \textbf{\textcolor{blue}{10.2}} & 93.6 & RTX 4090 \\
\Xhline{1pt}
\end{tabular}
\end{adjustbox}
\begin{tablenotes}
    \footnotesize
    \item Note: Since the receptive field of the YOLO-LLTS model adjusts according to the target size, both the number of parameters and FPS vary with changes in the dataset. Due to the lack of open-source code for MIAF-net, the experimental results are based on our own reproduction.
\end{tablenotes}
\end{table*}

\begin{table*}[htb]
\caption{Performance comparison on \textbf{GTSDB-night} dataset; The first and second best results are indicated in \textbf{\textcolor{blue}{blue}} and \textbf{\textcolor{green}{green}}}.
\label{gtsdb-night}
\renewcommand{\arraystretch}{1.5}
\begin{adjustbox}{width=\textwidth}
\begin{tabular}{cccccccccc>{\columncolor[HTML]{EFEFEF}}c}
\Xhline{1pt}
        Method & Venue & Input Size & Precision(\%) & Recall(\%) & F1 & mAP50(\%) & mAP50:95(\%) & Param(M)$\downarrow$ & FPS$\uparrow$ & GPU \\
\hline
        YOLOv5-L\cite{yolov5} & 2020 & 640$\times$640 & 52.8 & 55.3 & 54.0 & 54.4 & 45.3 & 46.1  & 97.1 & RTX 4090 \\
        YOLOv6-L\cite{yolov6} & CVPR2022 & 640$\times$640 & 71.4 & 45.4 & 55.5 & 56.2 & 48.3 & 59.5 & 75.1 & RTX 4090 \\
        GOLD-YOLO-L\cite{gold-yolo} & NeurIPS2023 & 640$\times$640 & 58.6 & 49.5 & 53.7 & 53.3 & 44.3 & 75.0 & 62.2 & RTX 4090 \\
        YOLOv8-L\cite{yolov8_ultralytics} & 2023 & 640$\times$640 & 71.3 & 50.2 & 58.9 & 61.6 & 52.5 & 43.6 & \textbf{\textcolor{green}{116.3}} & RTX 4090 \\
        YOLOv9-C \cite{yolov9} & CVPR2024 & 640$\times$640 & 54.8 & 52.4 & 53.6 & 53.3 & 46.1 & 50.7 & 88.5 & RTX 4090 \\
        YOLOv10-L\cite{yolov10} & NeurIPS2024 & 640$\times$640 & 66.9 	& 46.8 & 55.1 & 55.1 & 48.3 & 25.7 & 69.9 & RTX 4090 \\
        YOLOv11-L\cite{yolo11_ultralytics} & 2024 & 640$\times$640 & 58.8 & 50.7 & 54.5 & 56.7 & 49.0 & 25.3 & 72.5 & RTX 4090 \\
        YOLO-TS\cite{25} & arXiv2024 & 640$\times$640 & \textbf{\textcolor{green}{72.4}} & \textbf{\textcolor{green}{59.2}} & \textbf{\textcolor{green}{65.1}} & \textbf{\textcolor{green}{67.3}} & \textbf{\textcolor{green}{56.0}} & \textbf{\textcolor{green}{12.5}} & \textbf{\textcolor{blue}{149.3}} & RTX 4090 \\
        YOLOv12-L\cite{yolov12} & arXiv2025 & 640$\times$640 & 61.4 & 45.1 & 52.0 & 55.0 & 47.5 & 26.4 & 57.1 & RTX 4090 \\
\hline
        YOLO-LLTS(ours) & - & 640$\times$640 & \textbf{\textcolor{blue}{77.1(+4.7)$\uparrow$}} & \textbf{\textcolor{blue}{62.4(+3.2)$\uparrow$}} & \textbf{\textcolor{blue}{69.0(+3.9)$\uparrow$}} & \textbf{\textcolor{blue}{74.8(+7.5)$\uparrow$}} & \textbf{\textcolor{blue}{65.8(+9.8)$\uparrow$}} & \textbf{\textcolor{blue}{9.9}} & 101.0 & RTX 4090 \\
\Xhline{1pt}
\end{tabular}
\end{adjustbox}
\end{table*}

\begin{figure*}[htb]
    \centering
    \includegraphics[width=\textwidth]{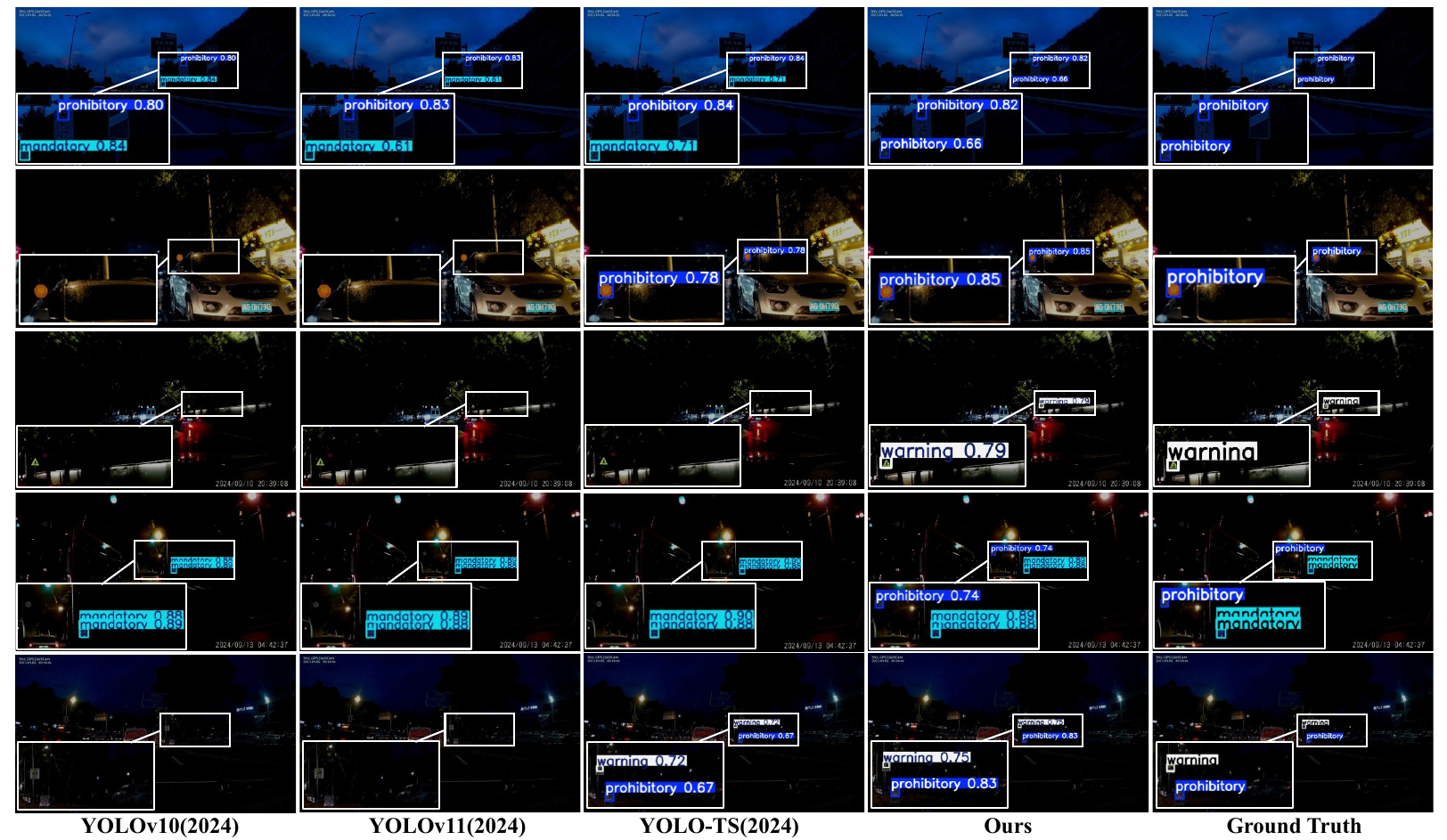}
    \caption{Comparison experimental results on the CNTSSS dataset are shown in the images. The first to third columns represent the latest advanced algorithms, the fourth column shows our model, and the fifth column represents the ground truth. Our detection results are consistent with the ground truth.}
    \label{compare}
\end{figure*}

As shown in Tab. \ref{tt100k}, although YOLO-TS demonstrates exceptionally high performance on the TT100K-night dataset, our model still outperforms YOLO-TS. Specifically, our model achieves an accuracy of 77.2\%, surpassing the second-ranked YOLO-TS by 2.0\%. In terms of recall, our model reaches 64.4\%, 1.2\% higher than YOLO-TS. Furthermore, our model achieves an F1 score of 70.2, surpassing YOLO-TS by 1.5. In the mAP50 and mAP50-95 evaluation metrics, which measure model detection performance at different IoU thresholds, our model also performs the best, achieving 71.4\% and 50.0\%, respectively. With a parameter count of 9.9M and an FPS of 83.3, our model maintains high efficiency in processing speed. Due to the lack of open-source code for MIAF-net, the experimental results are based on our own reproduction. Zhang et al. only released the trained weights for the CCTSDB2021 dataset. Therefore, experimental results for datasets other than CCTSDB2021 were obtained by training the model using the open-source code. The model's poor performance on the TT100K-night dataset may be attributed to the fact that TT100K-night is generated from daytime images, and the model does not generalize well to such generated data.

Tab. \ref{cntsss} presents a performance comparison of different traffic sign detection models on the CNTSSS dataset. Our model achieves an accuracy of 88.3\%, surpassing the second-ranked YOLO-TS by 1.0\%. In terms of recall, our approach reaches 74.9\%, which is 1.0\% higher than the second-ranked GOLD-YOLO-L, demonstrating superior detection coverage. Our model’s F1 score is 81.0, outperforming GOLD-YOLO-L by 1.5. The mAP50 of our model is 81.2\%, surpassing GOLD-YOLO-L by 1.3\%, and the mAP50-95 is 60.1\%, which is 1.9\% higher than YOLO-TS. With a parameter count of 13.9M and an FPS of 82.0, these results indicate that our approach maintains high accuracy and robustness across different detection difficulties.

Tab. \ref{cctsdb} presents a performance comparison of different traffic sign detection models on the CCTSDB2021 dataset. Despite the fact that the nighttime data used for training account for only a small portion of the dataset, with the majority consisting of daytime traffic sign data, our model still achieves outstanding results across multiple metrics. Specifically, our model reaches an accuracy of 88.8\%, surpassing the second-ranked YOLO-TS by 0.7\%. In terms of recall, our model achieves 81.1\%, surpassing the second-ranked YOLO-TS by 0.3\%. With an F1 score of 84.8, our model outperforms YOLO-TS by 0.5, reflecting a good balance between precision and recall. Our method also excels in the mAP50 and mAP50-95 metrics, which evaluate the model’s performance across different IoU thresholds, achieving 87.8\% and 57.5\%, surpassing YOLO-TS by 1.8\% and 0.3\%. With a parameter count of 10.2M and an FPS of 93.6, our model not only exceeds existing state-of-the-art models in accuracy but also demonstrates competitive processing speed, enabling rapid and accurate detection of traffic signs in practical applications. To rigorously assess whether the improvement is statistically significant, we trained each model 5 times with different random seeds and evaluated them on CCTSDB2021. The experiments show that the true value fluctuations for core metrics (Precision, Recall, F1, mAP50, mAP50:95) are within $\pm$0.4\%, indicating that the data exhibits small and stable variations. If the difference between the optimal and suboptimal results is less than or equal to 0.4\%, they can be considered statistically indistinguishable. We have marked such results with a \text{\textdagger} symbol in the top right corner.

Tab. \ref{gtsdb-night} presents a performance comparison of different traffic sign detection models on the GTSDB-night dataset. Our model achieves an accuracy of 77.1\%, surpassing the second-ranked YOLO-TS by 4.7\%. In terms of recall,  YOLO-LLTS reaches 62.4\%, which is 3.2\% higher than YOLO-TS, demonstrating superior detection coverage. Our model’s F1 score is 69.0, outperforming YOLO-TS by 3.9. Our method also excels in the mAP50 and mAP50-95 metrics, achieving 74.8\% and 65.8\%, surpassing YOLO-TS by 7.5\% and 9.8\%. With a parameter count of 9.9M and an FPS of 101.0, these results indicate that our approach maintains high accuracy and robustness across different detection difficulties. The excellent performance on the GTSDB-night dataset demonstrates the applicability of our model to different regions or countries with varying traffic sign styles.

As shown in Fig. \ref{compare}, a comparison is conducted on the CNTSSS dataset with the latest YOLO versions, including YOLOv10 \cite{yolov10}, YOLOv11 \cite{yolo11_ultralytics}, and YOLO-TS \cite{25}, released in 2024. The first row of images shows that, except for our model, all other models exhibit false detections. From the second to the fifth row, several models demonstrate instances of missed detections. In contrast, our model's results align perfectly with the ground truth, accurately recognizing traffic signs.

Our model has been trained and tested on both Chinese and German datasets. Experiments show that our model demonstrates a certain degree of generalization across different countries and regions. The design of our model is not dependent on the specific traffic sign styles of any particular country or region. The PGFE module, based on image enhancement and feature enhancement principles, is applicable to any low-light image. The anchor size of HRFM-SOD can be dynamically adjusted according to the target size, supporting sign proportions from different countries. When deploying the model to other countries or regions, it is recommended to fine-tune or retrain with local public datasets to further eliminate biases.

\begin{table}[t]
\renewcommand{\arraystretch}{1.5}
\caption{Ablation study for different components on CNTSSS. The best result is marked in \textbf{\textcolor{blue}{blue}}.}
\centering
\label{table_ablation_1}
\resizebox{0.49\textwidth}{!}{
\begin{tabular}{c|ccc|ccc }
\Xhline{1pt}
Model                          & HRFM-SOD & PGFE & MFIA & mAP50 &mAP50:95 &FPS$\uparrow$ \\ \hline
Baseline        &      &       &      & 75.1                 & 53.3      & 100.0                   \\
                & \checkmark    &       &      & 77.6                & 55.5      & \textbf{\textcolor{blue}{112.4}}                    \\
                &      &\checkmark       &      & 78.3                & 55.5      & 74.1                    \\                
                & \checkmark    & \checkmark     &      & 79.5                 & 59.1      & 78.7                    \\
                & \checkmark    &      &\checkmark      & 79.6                 & 58.4      & 102.0                    \\
\textbf{Ours}   & \checkmark    & \checkmark     & \checkmark    & \textbf{\textcolor{blue}{81.2$\uparrow$}}        & \textbf{\textcolor{blue}{60.1$\uparrow$}}      & 82.0  \\
\Xhline{1pt}
\end{tabular}
}
\end{table}

\subsection{Ablation Study}

\begin{figure*}[t]
    \centering
    \includegraphics[width=\textwidth]{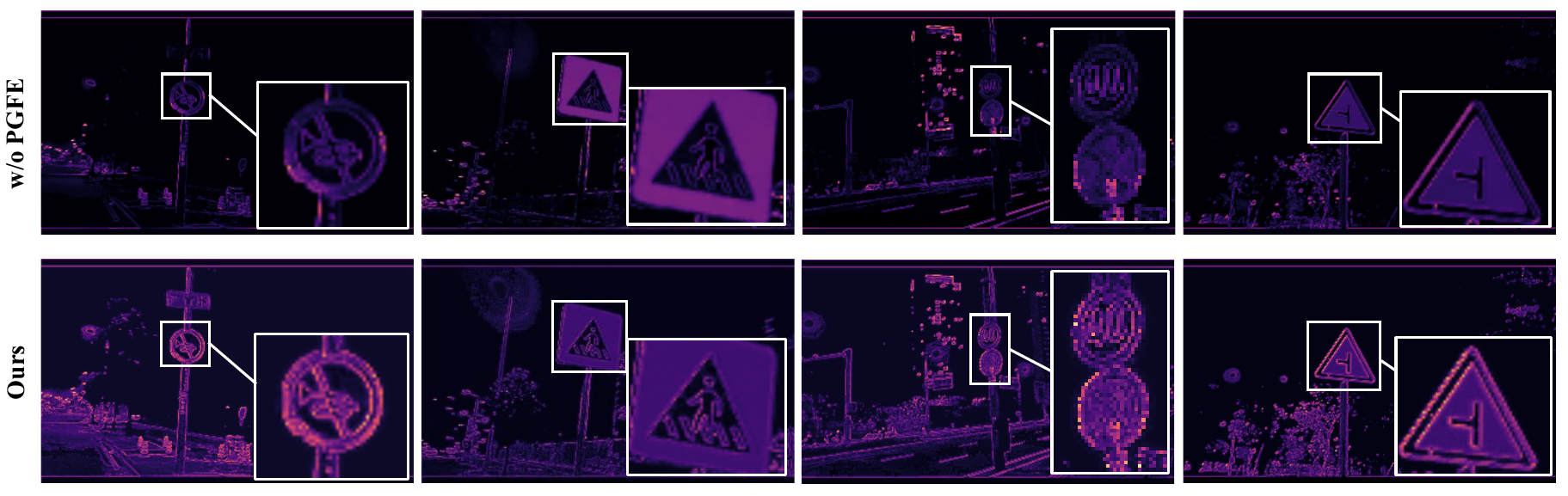}
    \caption{Comparisons of feature visualizations with and without the PGFE module.}
    \label{wo_pgfe}
\end{figure*}

In this section, ablation experiments are set to verify the rationality of the different modules. To thoroughly evaluate the effectiveness of traffic sign recognition under low-light conditions, we perform comparisons using the mAP50 and mAP50:95 metrics on the CNTSSS dataset.
\subsubsection{Effectiveness of HRFM-SOD}To validate the effectiveness of HRFM-SOD, we retain only this module on top of the baseline. As shown in Tab. \ref{table_ablation_1}, the HRFM-SOD module improves the mAP50 from 75.1\% to 77.6\%, an increase of 2.5\%, and the mAP50:95 from 53.3\% to 55.5\%, an increase of 2.2\%. The FPS also increases from 75.4 to 77.0. The experiment demonstrates that the HRFM-SOD module not only enhances detection performance but also accelerates model inference speed.

\newcolumntype{L}{>{\columncolor{gray!10}}c}
\begin{table}[htb]
\centering
\caption{Performance of \textbf{mAP50} on the CNTSSS dataset for \\different $\gamma$ and $\delta$. The best result is marked in \textbf{\textcolor{blue}{blue}}.}
\label{table_ablation_2_1}
\renewcommand{\arraystretch}{1.5}
\setlength{\tabcolsep}{10pt}
\resizebox{0.47\textwidth}{!}{
\begin{tabular}{L|ccccc}
\Xhline{1pt}
\rowcolor{gray!10}
\(\gamma\!\setminus\!\delta\) & 1.0 & 1.5 & 2.0 & 2.5 & 3.0\\
\hline
1.0 & 79.7 & 79.6 & 78.1 & 79.9 & 79.8\\
1.5 & 80.3 & 79.6 & 80.3 & 80.0 & 79.1\\
2.0 & 80.1 & 80.0 & 80.7 & \textcolor{blue}{\textbf{81.2$\uparrow$}} & 80.0\\
2.5 & 80.0 & 80.2 & 80.5 & 80.6 & 79.7\\
3.0 & 79.2 & 79.3 & 79.1 & 78.7 & 80.3\\
\Xhline{1pt}
\end{tabular}
}
\end{table}

\newcolumntype{L}{>{\columncolor{gray!10}}c}
\begin{table}[htb]
\centering
\caption{Performance of \textbf{mAP50:95} on the CNTSSS dataset for different $\gamma$ and $\delta$. The best result is marked in \textbf{\textcolor{blue}{blue}}.}
\label{table_ablation_2_2}
\renewcommand{\arraystretch}{1.5}
\setlength{\tabcolsep}{10pt}
\resizebox{0.47\textwidth}{!}{
\begin{tabular}{L|ccccc}
\Xhline{1pt}
\rowcolor{gray!10}
\(\gamma\!\setminus\!\delta\) & 1.0 & 1.5 & 2.0 & 2.5 & 3.0\\
\hline
    1.0 & 58.8 & 58.5 & 57.1 & 58.7 & 58.8\\
    1.5 & 59.2 & 58.6 & 58.5 & 58.5 & 57.5\\
    2.0 & 58.6 & 58.7 & 59.3 & \textbf{\textcolor{blue}{60.1$\uparrow$}} & 58.8\\
    2.5 & 58.7 & 58.5 & 58.9 & 58.1 & 57.6\\
    3.0 & 58.2 & 57.7 & 57.5 & 57.6 & 58.6\\
\Xhline{1pt}
\end{tabular}
}
\end{table}

\subsubsection{Effectiveness of PGFE} To validate the effectiveness of PGFE, we conducted corresponding ablation experiments. As shown in Tab. \ref{table_ablation_1}, the PGFE module improves the mAP50 from 75.1\% to 78.3\%, an increase of 3.2\%, and the mAP50:95 from 53.3\% to 55.5\%, an increase of 2.2\%. When PGFE and HRFM-SOD are combined, the mAP50 increases by 4.4\%, and the mAP50:95 increases by 5.8\%. The experiments demonstrate that the PGFE module plays a positive role in improving the model's detection accuracy and makes the largest contribution to accuracy improvement among several modules. Although the FPS decreased after introducing PGFE, from 100.0 in the baseline to 74.1, it still meets the requirements for autonomous driving and driver assistance systems. Therefore, PGFE achieves significant nighttime detection gains at an acceptable computational cost, making it valuable for practical deployment.

PGFE is not a traditional pixel-level image enhancement network but a feature-level enhancement module. Therefore, evaluating the enhancement performance of PGFE cannot rely on traditional image enhancement metrics. We output features with and without PGFE, and visualized the 64-channel feature map. As shown in Fig. \ref{wo_pgfe}, the feature map with PGFE clearly shows sharper edges, better contrast between light and dark, and some contours that were hidden in the darkness becoming visible. This experiment further proves that PGFE can improve image quality. Additionally, we conducted experiments with different parameter settings in the formula. As shown in Tab. \ref{table_ablation_2_1} and Tab. \ref{table_ablation_2_2}, the model performs best when $\gamma$ and $\delta$ are set to 2 and 2.5, respectively. The experimental range of values is set empirically to be between 1 and 3.
\subsubsection{Effectiveness of MFIA}Since the MFIA module can only be used in conjunction with the HRFM-SOD module, we validated the combined effect of both modules. As shown in the second to last row of Tab. \ref{table_ablation_1}, the mAP50 and mAP50:95 increased by 4.5\% and 5.1\%, respectively, compared to the baseline. When compared to the HRFM-SOD-only configuration, mAP50 and mAP50:95 increased by 2.0\% and 2.9\%, respectively. The experiment demonstrates that the MFIA module effectively integrates multi-receptive field features, improving detection performance.

\begin{figure*}[t]
    \centering
    \includegraphics[width=\textwidth]{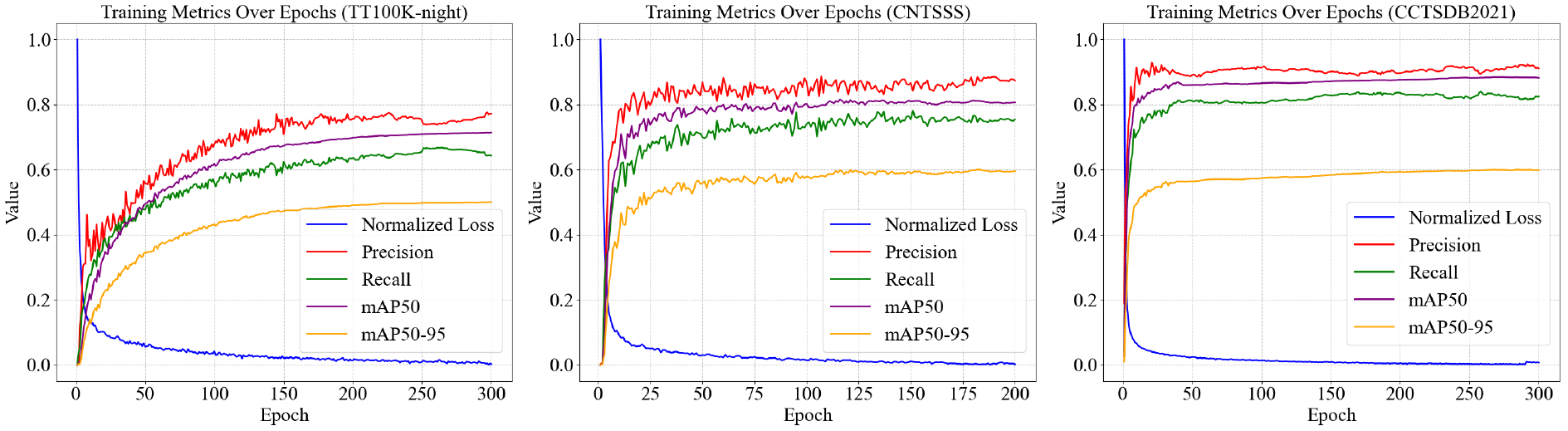}
    \caption{An analysis of several critical performance metrics of our model during the training phase, evaluated on the TT100K-night, CNTSSS, and CCTSDB2021 datasets.}
    \label{line}
\end{figure*}

\begin{figure}[t]
    \centering
    \includegraphics[width=0.48\textwidth]{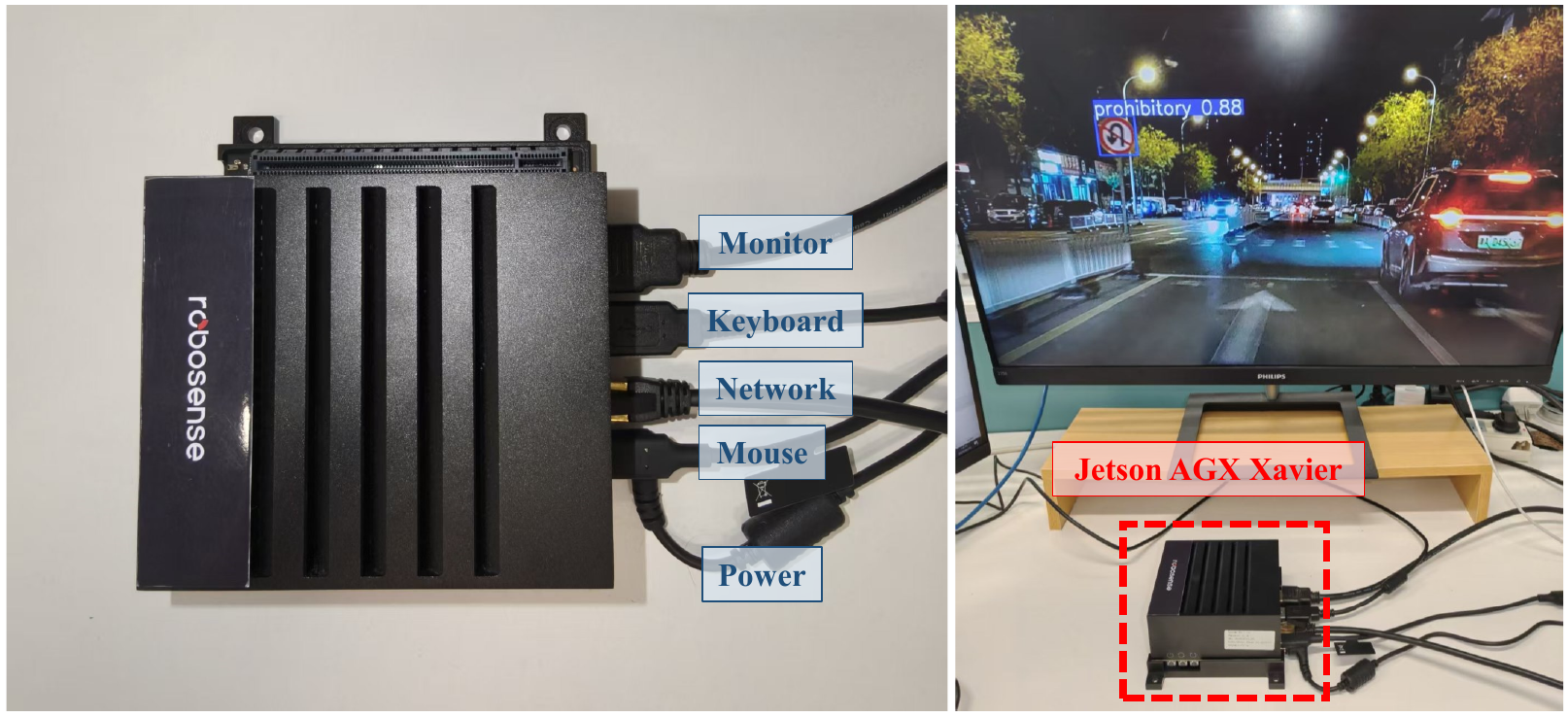} % 调整宽度为栏宽的一半左右
    \caption{Mobile Edge Computing Device NVIDIA Jetson AGX Xavier.}
    \label{Xavier}
\end{figure}

\begin{figure}[t]
    \centering
    \includegraphics[width=0.37\textwidth]{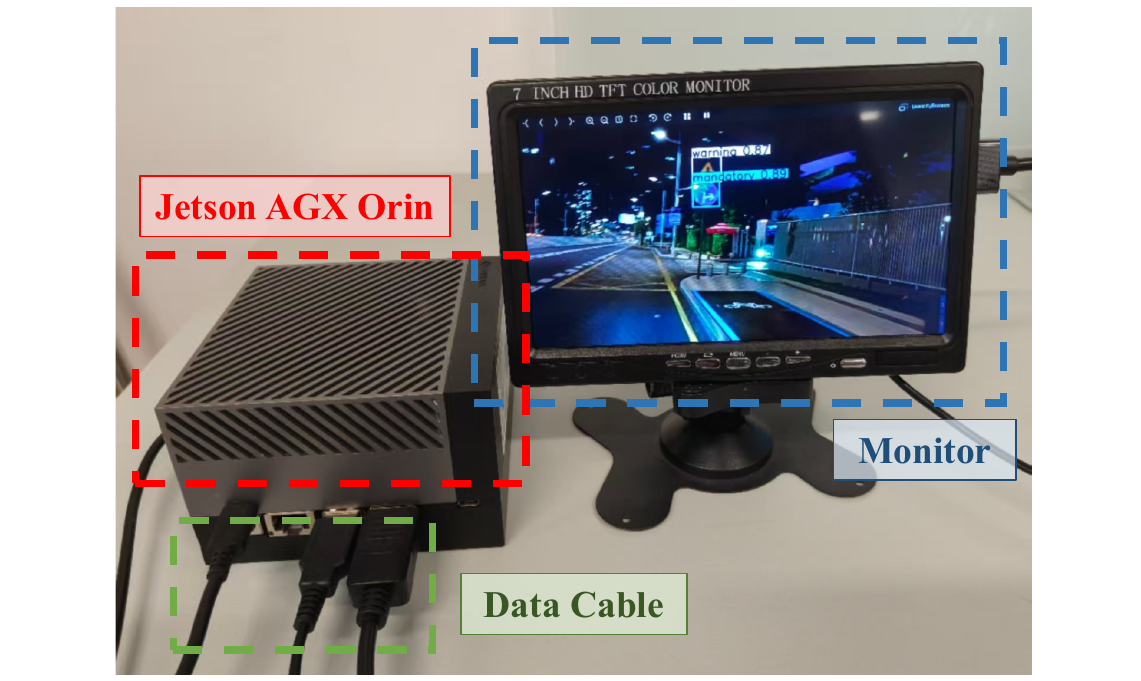} % 调整宽度为栏宽的一半左右
    \caption{Mobile Edge Computing Device NVIDIA Jetson AGX Orin.}
    \label{Orin}
\end{figure}

\subsection{Error Analysis and Model Deployment}
In this study, we conducted a detailed analysis of the training errors of the traffic sign detection model across three different datasets \cite{error}. Fig. \ref{line} presents the evolution of key training metrics, including normalized loss, precision, recall, mAP50, and mAP50-95, with respect to the number of training epochs. The normalized loss is the sum of the box loss, classification loss, and distribution focal Loss, which we have normalized for ease of presentation. As shown in Fig. \ref{line}, the normalized loss decreases rapidly and eventually stabilizes as the number of training epochs increases across all datasets, indicating that the model exhibits good convergence on these datasets. These critical performance metrics quickly increase in the early stages of training and eventually stabilize.

At the same time, we test the detection accuracy and inference times on mobile edge devices, NVIDIA Jetson AGX Xavier and NVIDIA Jetson AGX Orin. As shown in Fig. \ref{Xavier} and Fig. \ref{Orin}, we first connect the device to the display, keyboard, and mouse, then connect the network cable and power supply. We power on the edge computing device and test the trained weights on both the Jetson AGX Xavier and Jetson AGX Orin. The experiments on Jetson AGX Xavier were conducted in an Ubuntu 20.04 operating system environment, with optimization support provided by Jetpack 5.1. The experiments on Jetson AGX Orin were conducted in an Ubuntu 18.04 operating system environment, using the PyTorch deep learning framework (version 2.1.0), with optimization support provided by Jetpack 5.1. We selected 500 nighttime images from the CCTSDB2021 dataset for testing. As shown in Tab. \ref{device_X} and Tab. \ref{device_O}, YOLO-LLTS achieves the best performance in terms of accuracy. Specifically, our model outperforms the average of YOLOv8 to YOLOv12 by 4.6 in accuracy, 6.7 in recall, and 5.8 in F1 score. It also exceeds the average by 7.4 in mAP50 and 4.7 in mAP50:95. On edge devices, the peak memory usage for YOLO-LLTS inference on a single 640×640 image is 88 MB, which accounts for 18\% of the GPU reserved pool (88/490 MB), far from reaching the hardware limit. This result indicates that the model can still maintain usability in constrained edge environments. The number of parameters on the Xavier is 10.2 million, with an inference time of 138.8 ms. On the Orin, the number of parameters is 10.2 million, with an inference time of 44.9 ms. The experiment demonstrates that our model maintains relatively competitive performance even on devices with constrained power.

\begin{figure*}[t]
    \centering
    \includegraphics[width=\textwidth]{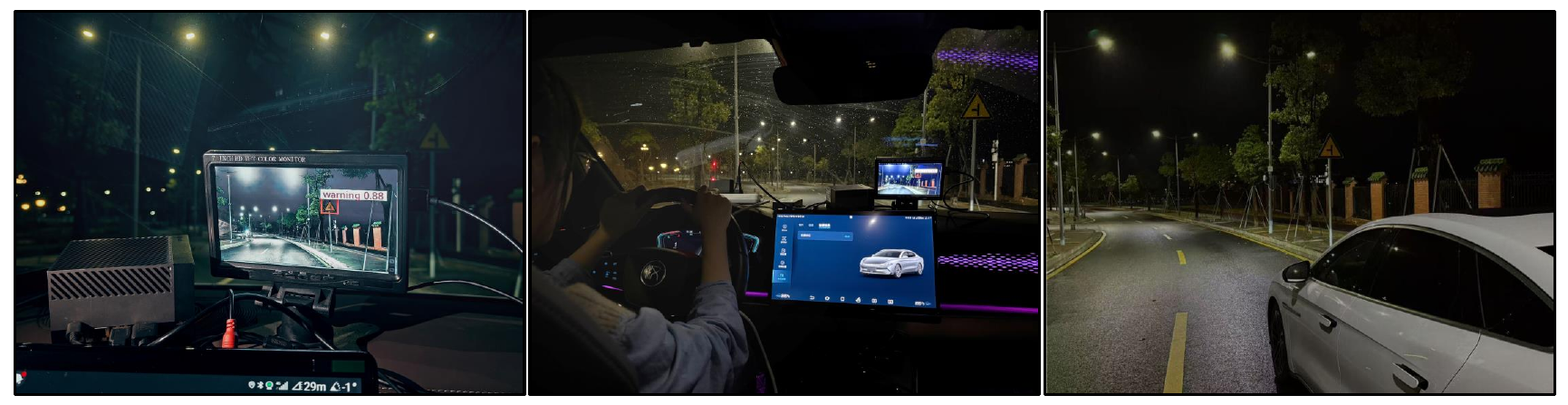}
    \caption{Conducting field tests using Mobile Edge Computing Devices.}
    \label{edge_em}
\end{figure*}

To evaluate the performance of our model in real-world scenarios, we use the mobile edge device to detect traffic signs on safe roads. We connect the Jetson AGX Orin to the vehicle's power supply, capture images of traffic signs using the camera, and then conduct testing using the trained weights. In the scene depicted in Fig. \ref{edge_em}, traffic signs were successfully detected and accurately classified. This demonstrates the full effectiveness of our model in real-world applications.

\begin{table*}[htb]
\caption{Testing the detection accuracy and inference times on \textbf{Jetson AGX Xavier}; The first best results are indicated in \textbf{\textcolor{blue}{blue}}.}
\label{device_X}
\renewcommand{\arraystretch}{1.5}
\begin{adjustbox}{width=\textwidth}
\begin{tabular}{ccccccccc>{\columncolor[HTML]{EFEFEF}}c}
\Xhline{1pt}
        Method & Input Size & Precision(\%) & Recall(\%) & F1 & mAP50(\%) & mAP50:95(\%) & Param(M)$\downarrow$ & Inference time (ms)$\downarrow$ & Device \\
\hline
        YOLOv8-L\cite{yolov8_ultralytics} & 640$\times$640 & 84.4 & 74.4 & 79.1 & 80.5 & 53.0 & 43.6 & 145.8 & AGX Xavier \\
        YOLOv9-C \cite{yolov9} & 640$\times$640 & 84.8 & 76.0 & 80.2 & 81.9 & 53.7 & 50.7 & 234.0 & AGX Xavier \\   
        YOLOv10-L\cite{yolov10} & 640$\times$640 & 83.9 & 73.2 & 78.2 & 78.7 & 51.4 & 25.7 & 134.2 & AGX Xavier \\
        YOLOv11-L\cite{yolo11_ultralytics} & 640$\times$640 & 85.8 & 74.2 & 79.6 & 81.2 & 54.1 & 25.3 & \textbf{\textcolor{blue}{110.9}} & AGX Xavier \\
        YOLOv12-L\cite{yolov12} & 640$\times$640 & 84.9 & 74.2 & 79.2 & 82.8 & 55.7 & 26.3 & 134.8 & AGX Xavier \\
        Average & - & 84.2 & 74.4 & 79.0 & 80.4 & 52.8 & 34.3 & 151.9 & - \\
\hline     
        YOLO-LLTS(ours) & 640$\times$640 & \textbf{\textcolor{blue}{88.8$\uparrow$}} & \textbf{\textcolor{blue}{81.1$\uparrow$}} & \textbf{\textcolor{blue}{84.8$\uparrow$}} & \textbf{\textcolor{blue}{87.8$\uparrow$}} & \textbf{\textcolor{blue}{57.5$\uparrow$}} & \textbf{\textcolor{blue}{10.2}} & 138.8 & AGX Xavier \\
\Xhline{1pt}
\end{tabular}
\end{adjustbox}
\end{table*}

\begin{table*}[htb]
\caption{Testing the detection accuracy and inference times on \textbf{Jetson AGX Orin}; The first best results are indicated in \textbf{\textcolor{blue}{blue}}.}
\label{device_O}
\renewcommand{\arraystretch}{1.5}
\begin{adjustbox}{width=\textwidth}
\begin{tabular}{ccccccccc>{\columncolor[HTML]{EFEFEF}}c}
\Xhline{1pt}
        Method & Input Size & Precision(\%) & Recall(\%) & F1 & mAP50(\%) & mAP50:95(\%) & Param(M)$\downarrow$ & Inference time (ms)$\downarrow$ & Device \\
\hline
        YOLOv8-L\cite{yolov8_ultralytics} & 640$\times$640 & 84.4 & 74.4 & 79.1 & 80.5 & 53.0 & 43.6 & \textbf{\textcolor{blue}{24.6}} & AGX Orin \\
        YOLOv9-C \cite{yolov9} & 640$\times$640 & 84.4 & 76.0 & 80.0 & 81.9 & 53.7 & 50.7 & 55.9 & AGX Orin \\    
        YOLOv10-L\cite{yolov10} & 640$\times$640 & 83.9 & 73.2 & 78.2 & 78.7 & 51.4 & 25.7 & 36.8 & AGX Orin \\
        YOLOv11-L\cite{yolo11_ultralytics} & 640$\times$640 & 85.8 & 74.2 & 79.6 & 81.2 & 54.1 & 25.3 & 34.7 & AGX Orin \\
        YOLOv12-L\cite{yolov12} & 640$\times$640 & 84.9 & 74.2 & 79.2 & 82.8 & 55.7 & 26.3 & 48.4 & AGX Orin \\
        Average & - & 84.1 & 74.4 & 79.0 & 80.4 & 52.8 & 34.3 & 27.0 & - \\
\hline
        YOLO-LLTS(ours) & 640$\times$640 & \textbf{\textcolor{blue}{88.8$\uparrow$}} & \textbf{\textcolor{blue}{81.1$\uparrow$}} & \textbf{\textcolor{blue}{84.8$\uparrow$}} & \textbf{\textcolor{blue}{87.8$\uparrow$}} & \textbf{\textcolor{blue}{57.5$\uparrow$}} & \textbf{\textcolor{blue}{10.2}} & 44.9 & AGX Orin \\
\Xhline{1pt}
\end{tabular}
\end{adjustbox}
\end{table*}

\begin{figure*}[t]
    \centering
    \includegraphics[width=\textwidth]{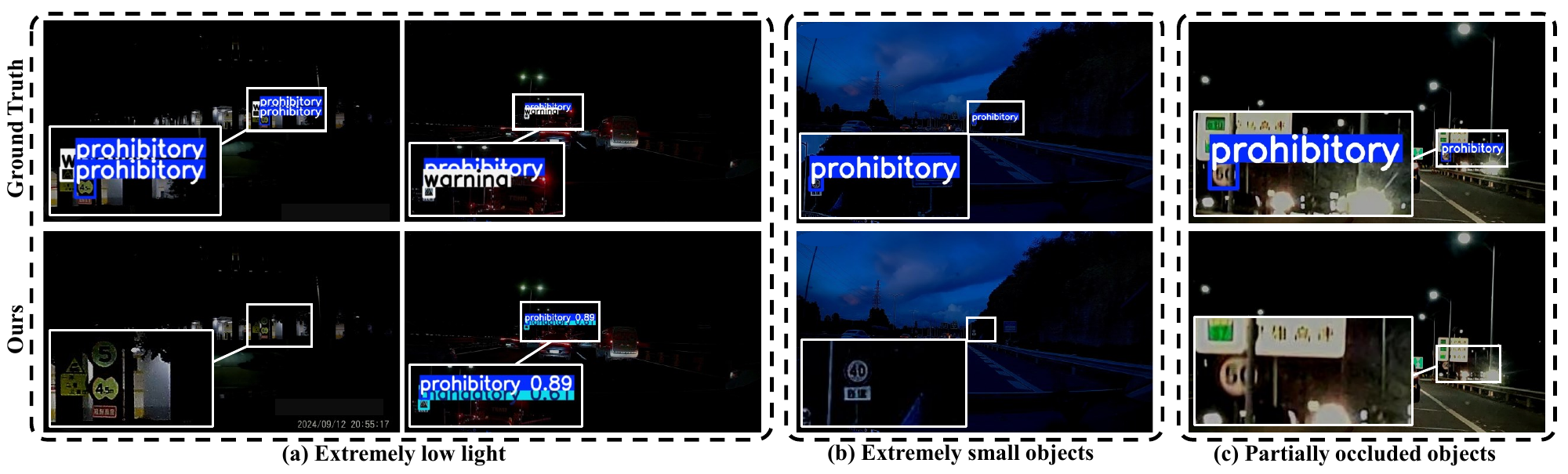}
    \caption{Some failure cases under extremely low light, extremely small objects, and partially occluded objects conditions.}
    \label{fail}
\end{figure*}

\section{Conclusion}
\label{sec:conclusion}
Under low-light conditions, the perception systems of autonomous vehicles are prone to false positives and missed detections, leading to road traffic accidents. In this paper, we propose YOLO-LLTS, an end-to-end real-time traffic sign detection algorithm, which effectively enhances the accuracy and speed of traffic sign recognition, thereby improving the safety of ADAS and autonomous vehicles in low-light environments. To address the lack of nighttime-specific traffic sign datasets, we constructed a novel dataset named Chinese Nighttime Traffic Sign Sample Set (CNTSSS). This dataset includes images captured under varying low-light conditions from dusk to midnight, covering diverse scenarios such as urban, highway, and rural environments under different weather conditions. We introduce the High-Resolution Feature Map for Small Object Detection (HRFM-SOD) module to effectively address the indistinct features of small objects under low-light conditions, significantly improving both detection accuracy and inference speed. Moreover, we design the Multibranch Feature Interaction Attention (MFIA) module, enabling deep interaction and fusion of features across multiple receptive fields, thereby enhancing the model's ability to capture and utilize critical information. Furthermore, we propose the Prior-Guided Feature Enhancement Module (PGFE) to alleviate challenges such as increased noise, reduced contrast, and blurriness under low-light environments, significantly boosting the detection performance. Experimental results demonstrate that our approach achieves state-of-the-art performance on the TT100K-night, CNTSSS, CCTSDB2021, and GTSDB-night datasets. The deployment experiments on edge devices further validate the practical effectiveness and real-time applicability of our method. 

Although the YOLO-LLTS algorithm shows significant improvements in traffic sign detection and recognition under low-light conditions, there are still some failure cases. As shown in Fig. \ref{fail}(a), when the road environment has weak lighting or lacks street lamps, our model encounters false positives and false negatives. As shown in Fig. \ref{fail}(b), when the target is extremely small, the model suffers from missed detections. As shown in Fig. \ref{fail}(c), when the sign is obstructed by barriers, the model fails to detect the target. Future research could further explore ways to enhance the model's detection performance under extremely low light, extremely small objects, and partially occluded objects conditions. Additionally, both our CNTSSS dataset and existing public datasets do not contain data on extreme weather conditions, such as heavy fog or heavy rain. Therefore, the model's effectiveness under extreme weather remains unverified, posing a risk of false negatives and false positives. In the future, we plan to expand the CNTSSS dataset to include data on extreme weather conditions and optimize the model to better adapt to challenging weather situations. Furthermore, we plan to expand the CNTSSS dataset to include more countries or regions, systematically assessing the model's robustness across different traffic regulatory environments. The code and the dataset are available at https://github.com/linzy88/YOLO-LLTS.

\section*{Acknowledgments}
This project is jointly supported by the National Natural Science Foundation of China (Nos.52172350, W2421069 and 51775565), the Tongchuang Intelligent Medical Inter-disciplinary Talent Training Fund of Sun Yat-sen University (No.76160-54990001), the Guangdong Basic and Applied Research Foundation (No. 2022B1515120072), the Guangzhou Science and Technology Plan Project (No.2024B01W0079), the Nansha Key RD Program (No.2022ZD014), the Science and Technology Planning Project of Guangdong Province (No.2023B1212060029). We thank Dr. Dezong Zhao from the University of Glasgow and Dr. Kening Li from Shenzhen Institute of Information Technology for their valuable comments.

%{\appendix[Proof of the Zonklar Equations]
%Use $\backslash${\tt{appendix}} if you have a single appendix:
%Do not use $\backslash${\tt{section}} anymore after $\backslash${\tt{appendix}}, only $\backslash${\tt{section*}}.
%If you have multiple appendixes use $\backslash${\tt{appendices}} then use $\backslash${\tt{section}} to start each appendix.
%You must declare a $\backslash${\tt{section}} before using any $\backslash${\tt{subsection}} or using $\backslash${\tt{label}} ($\backslash${\tt{appendices}} by itself
% starts a section numbered zero.)}

%{\appendices
%\section*{Proof of the First Zonklar Equation}
%Appendix one text goes here.
% You can choose not to have a title for an appendix if you want by leaving the argument blank
%\section*{Proof of the Second Zonklar Equation}
%Appendix two text goes here.}

\bibliographystyle{IEEEtran}
\bibliography{reference}

% \enlargethispage{-80mm}
\begin{IEEEbiography}[{\includegraphics[width=1in,height=1.25in,clip,keepaspectratio]{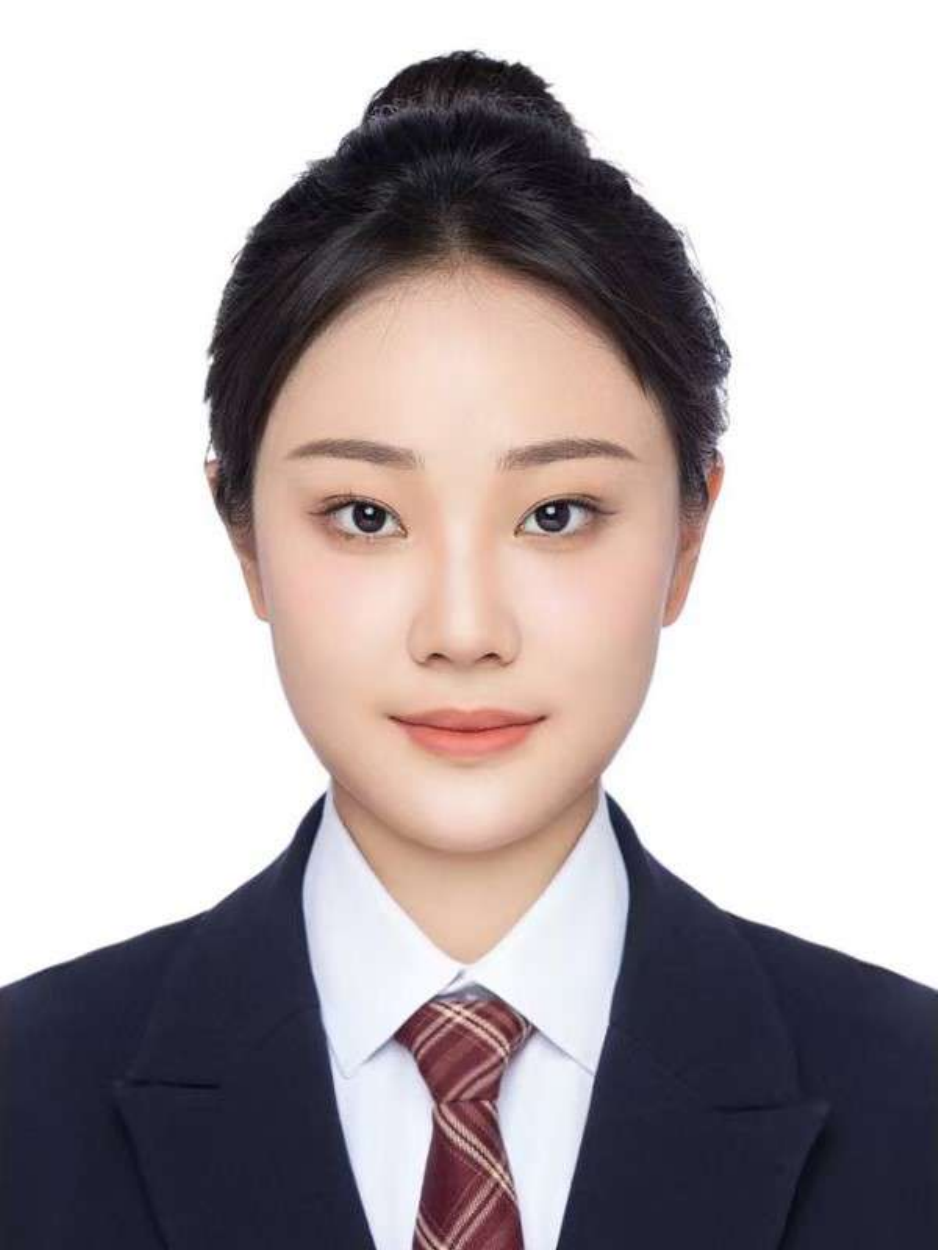}}]{Ziyu Lin} is currently pursuing a B.Sc. degree in Traffic Engineering with the School of intelligent systems engineering, Sun Yat-sen University, Guangzhou, China. Her research interests include computer vision, deep learning, and autonomous driving technology, with a particular focus on exploring how these technologies can enhance the efficiency and safety of autonomous driving.
\end{IEEEbiography}
\vspace{-1.6cm}
\begin{IEEEbiography}[{\includegraphics[width=1in,height=1.25in,clip,keepaspectratio]{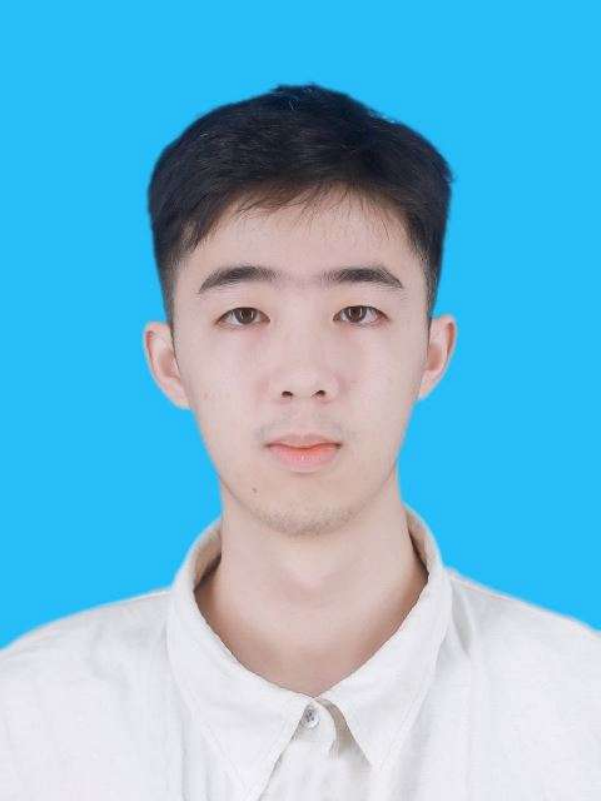}}]{Yunfan Wu} is currently pursuing a B.Sc. degree in Traffic Engineering with the School of intelligent systems engineering, Sun Yat-sen University, Guangzhou, China. His research interests include autonomous vehicles, computer vision and deep learning, with a commitment to reducing the likelihood of accidents in autonomous vehicles and enhancing their operational efficiency.
\end{IEEEbiography}
\vspace{-1.6cm}
\begin{IEEEbiography}[{\includegraphics[width=1in,height=1.25in,clip,keepaspectratio]{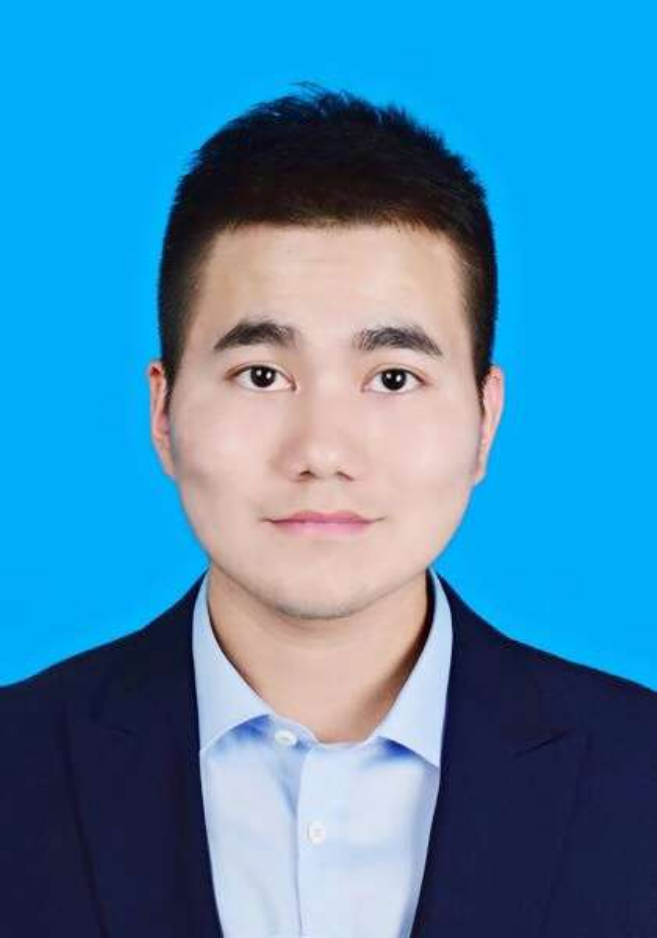}}]{Yuhang Ma} received his B.S. degree in Automation at Huazhong University of Science and Technology in 2020. He is currently pursuing the master’s degree in Electronic and Information Engineering at Sun Yat-sen University, Shenzhen, 518107, Guangdong, P.R.China. His current research interests include autonomous driving and computer vision.
\end{IEEEbiography}
% \vspace{-1.6cm}
\newpage
\begin{IEEEbiography}[{\includegraphics[width=1in,height=1.25in,clip,keepaspectratio]{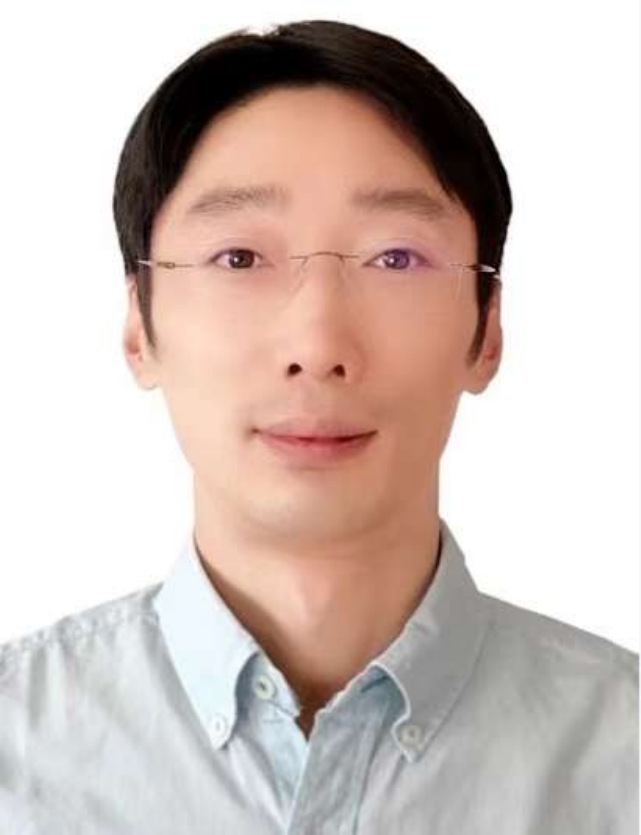}}]{Junzhou Chen} received his Ph.D. in Computer Science and Engineering from the Chinese University of Hong Kong in 2008, following his M.Eng degree in Software Engineering and B.S. in Computer Science and Applications from Sichuan University in 2005 and 2002, respectively. Between March 2009 and February 2019, he served as a Lecturer and later as an Associate Professor at the School of Information Science and Technology at Southwest Jiaotong University. He is currently an associate professor at the Guangdong Provincial Key Laboratory of Intelligent Transportation System, School of Intelligent Systems Engineering, Sun Yat-Sen University, Guangzhou 510275, China. His research interests include computer vision, machine learning, intelligent transportation systems, mobile computing and medical image processing.
\end{IEEEbiography}
\vspace{-1.6cm}
% \newpage
\begin{IEEEbiography}[{\includegraphics[width=1in,height=1.25in,clip,keepaspectratio]{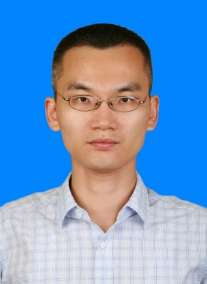}}]{Ronghui Zhang} received a B.Sc. (Eng.) from the Department of Automation Science and Electrical Engineering, Hebei University, Baoding, China, in 2003, an M.S. degree in Vehicle Application Engineering from Jilin University, Changchun, China, in 2006, and a Ph.D. (Eng.) in Mechanical \& Electrical Engineering from Changchun Institute of Optics, Fine Mechanics and Physics, the Chinese Academy of Sciences, Changchun, China, in 2009. After finishing his post-doctoral research work at INRIA, Paris, France, in February 2011, he is currently an Associate Professor with the Guangdong Provincial Key Laboratory of Intelligent Transportation System, School of Intelligent Systems Engineering, Sun Yat-Sen University, Guangzhou 510275, China. His current research interests include computer vision, intelligent control and ITS.
\end{IEEEbiography}
\vspace{-1.6cm}
\begin{IEEEbiography}[{\includegraphics[width=1in,height=1.25in,clip,keepaspectratio]{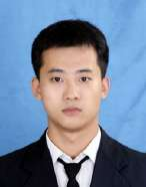}}]{Jiaming Wu} received the B.S. and M.S. degree in School of Transportation Science and Engineering from Harbin Institute of Technology in 2014, and Ph.D. degree from Southeast University in 2019. He is currently an Assistant Professor in the Department of Architecture and Civil Engineering, Chalmers University of Technology, Gothenburg, Sweden. His research interests include electric vehicle fleet management (routing and charging), connected and automated vehicles, and intersection control.
\end{IEEEbiography}
\vspace{-1.6cm}
\begin{IEEEbiography}[{\includegraphics[width=1in,height=1.25in,clip,keepaspectratio]{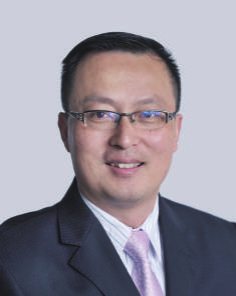}}]{Guodong Yin} received the Ph.D. degree in the Department of Vehicle Engineering, Southeast University, Nanjing, China, in 2007. From 2011 to 2012, he was a Visiting Scholar with the Department of Mechanical and Aerospace Engineering, The Ohio State University, Columbus, OH, USA. He is currently a Professor with the School of Mechanical Engineering, Southeast University, Nanjing, China.
His research interests include vehicle dynamics, connected vehicles, and multiagent control. He has published more than 150 peer-reviewed journal and conference papers. 
\end{IEEEbiography}
\vspace{-1.6cm}
\begin{IEEEbiography}[{\includegraphics[width=1in,height=1.25in,clip,keepaspectratio]{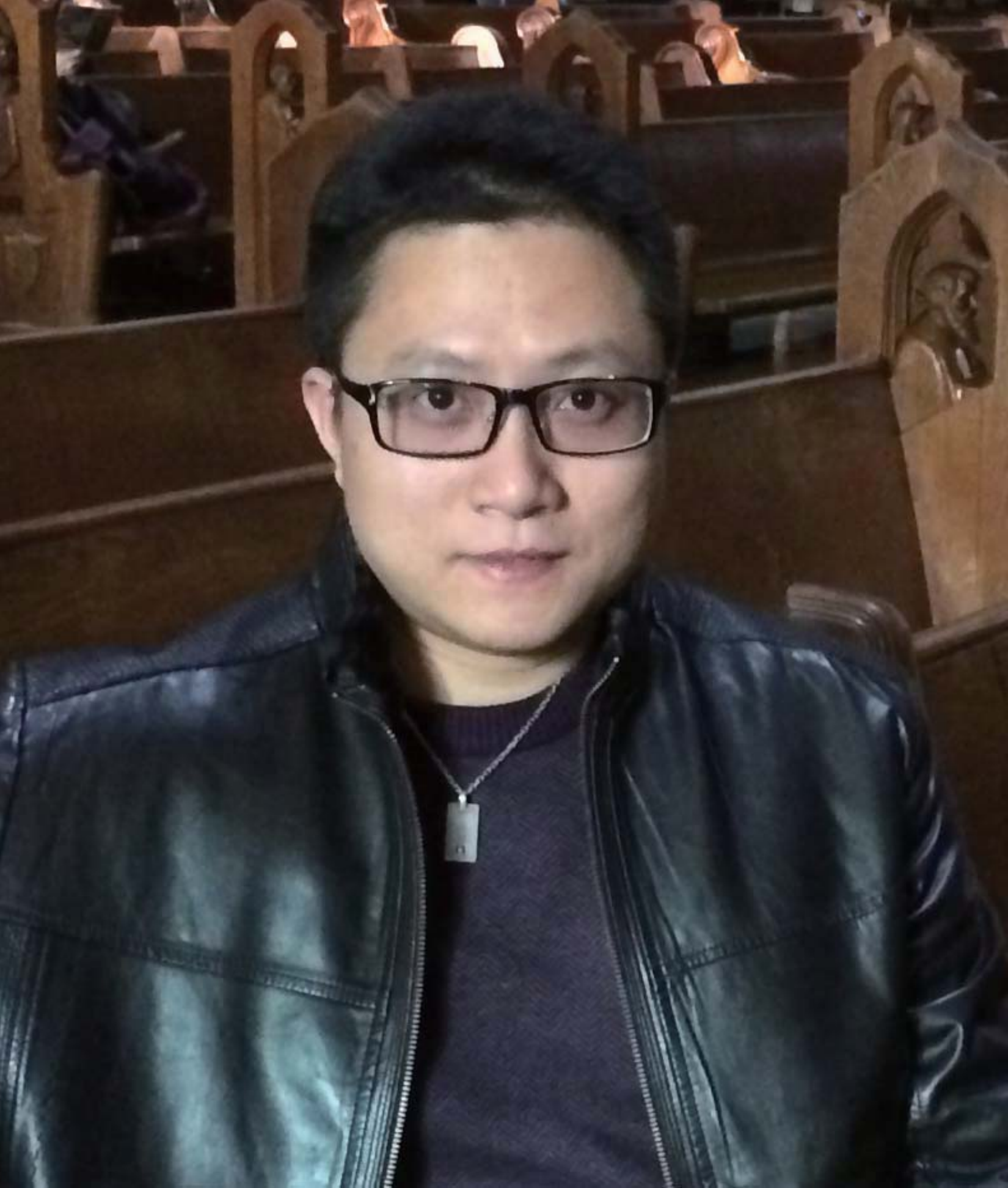}}]{Liang Lin} (Fellow, IEEE) is a full Professor at Sun Yat-sen University. He served as the Executive R\&D Director and Distinguished Scientist of SenseTime Group from 2016 to 2018, taking charge of transferring cutting-edge technology into products. He has authored or co-authored more than 200 papers in leading academic journals and conferences with more than 12,000 citations. He is an associate editor of IEEE Trans. Human-Machine Systems and IET Computer Vision. He served as Area Chairs for numerous conferences such as CVPR, ICCV, and IJCAI. He is the recipient of numerous awards and honors including Wu Wen-Jun Artificial Intelligence Award, CISG Science and Technology Award, ICCV Best Paper Nomination in 2019, Annual Best Paper Award by Pattern Recognition (Elsevier) in 2018, Best Paper Diamond Award in IEEE ICME 2017, Google Faculty Award in 2012, and Hong Kong Scholars Award in 2014. He is a Fellow of IAPR and IET. \end{IEEEbiography}

\end{document}